\documentclass{article}
\usepackage{iclr2027_conference,times}
\usepackage{graphicx,float}
\usepackage{microtype,url,hyperref}

\title{Task Context Shapes Component Roles\\in Language Models}
\author{Sovesh Mohapatra\textsuperscript{1}\qquad
Dani S Bassett\textsuperscript{1,2}\thanks{Corresponding author: \href{mailto:dani.bassett@yale.edu}{dani.bassett@yale.edu}.}\\
\textsuperscript{1}University of Pennsylvania\qquad
\textsuperscript{2}Yale University}

\iclrfinalcopy
\date{}
\hypersetup{
    hidelinks,
    pdftitle={Task Context Shapes Component Roles in Language Models},
    pdfauthor={Sovesh Mohapatra and Dani S Bassett}
}

\begin{document}
\maketitle
\lhead{}
\renewcommand{\headrulewidth}{0pt}

\begin{abstract}
Large language models reuse the same attention heads across tasks, but identifying these shared heads does not explain what they do differently in each task. A natural source of inspiration is task-dependent interactions among shared brain regions, which motivate a network view of language-model computation. Here, we compare fixed model components across tasks using identical graph facts, statements specifying which directed connections are present. We measure how interventions affect attention heads and the activation state in later layers (influence), as well as the answer (importance). Task changes redistribute the influence in five open-weight models spanning four families and 1.5–14 billion parameters. In Gemma-2-9B and Qwen3-14B, an activation edit changes an early internal activation vector and switches which fact the model reads while the written instruction stays fixed. The same heads then respond to different graph facts and influence later components differently. For each prompt, we construct two linear response maps, one before the edit and one after it. Each map predicts a later head’s response from a measured change in an earlier head. For a separate factual intervention, the map for the corresponding condition reduces mean squared prediction error by 53.6\% in Gemma and 5.0\% in Qwen compared with the average of the two maps. These findings show that task context changes both which graph facts the shared components respond to and how their outputs affect later computation. A network account may help explain how models use shared parameters to perform different computations. Measurements of task-dependent influence could also guide interventions intended to alter one behavior while preserving others.
\end{abstract}

\section{Introduction}

Large language models can perform different tasks with a single set of parameters. Instructions or a few in-context examples specify the task at inference time, without further parameter updates \citep{brown2020fewshot,wei2022finetuned}. Circuit analyses show that different tasks can reuse attention heads and substantial parts of the same circuit \citep{merullo2024circuit}. A small number of heads can also carry a compact representation of the task demonstrated in context \citep{todd2024function}. Two aspects of a shared head's computation may change across tasks: that is, which facts it responds to and how its output affects later components.

Network neuroscience provides a related perspective on how shared components support different tasks, focusing on interactions among brain regions \citep{bassett2017network}. The same regions can participate in different tasks, while the relationships between their activity change with the task \citep{cole2013multitask,krienen2014reconfigurable}. These relationships, known as functional connectivity, can help distinguish the task being performed \citep{gonzalezcastillo2015tracking}. In language models, functional networks have similarly been identified from groups of neurons that repeatedly become active together. Suppressing selected networks impairs performance, whereas amplifying their activity can improve performance on some tasks \citep{liu2025brainfunctional}. Tasks in the same cognitive domain, such as language or social reasoning, also share more task-relevant model neurons than tasks in different domains \citep{han2026modular}. Mechanistic circuit analyses complement this network perspective by examining how individual model components affect later computation and the answer.

Circuit discovery identifies components and interactions that support specific model behaviors \citep{wang2023ioi,conmy2023automated,marks2025sparse}. Some interactions between earlier and later heads can also be predicted from their weight matrices \citep{merullo2024talking}. Interventions on internal task representations can make a model perform a task on new inputs \citep{todd2024function}. Attention heads also help identify possible output labels \citep{yang2026localizing} and suppress information that is irrelevant to the task \citep{cho2026removal}. A central question remains: when the graph facts are held fixed, does changing the task alter which facts the same components respond to and how their outputs influence later components?

\begin{figure}[H]
    \centering
    \includegraphics[width=\linewidth]{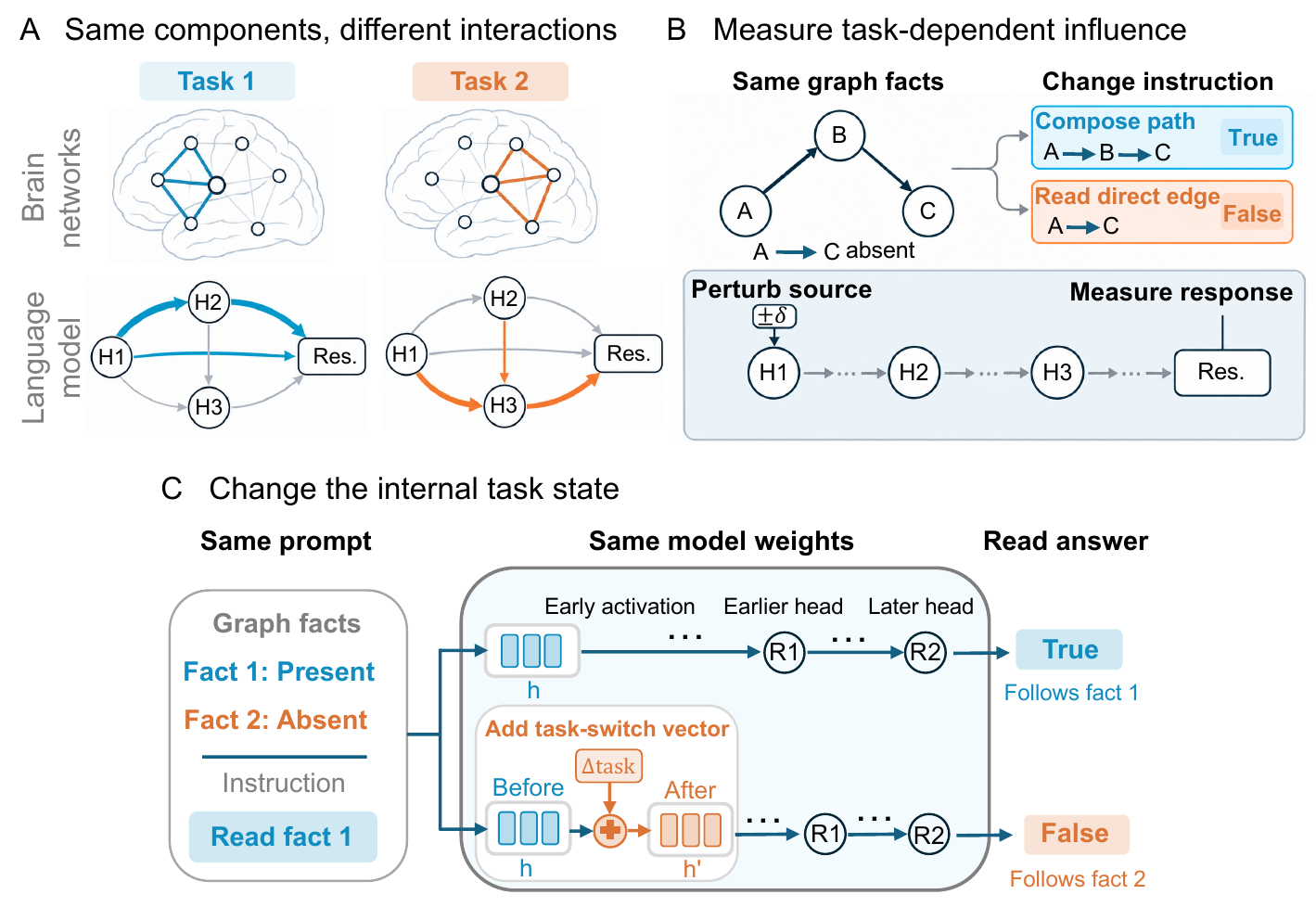}
    \caption{\textbf{Task-dependent computation with shared components.} \textbf{A. Same components, different interactions.} Brain networks motivate the hypothesis that shared model components support different tasks through changes in their interactions. H1--H3 denote fixed attention heads in layer order; Res.\ denotes a later residual-stream site. \textbf{B. Measure task-dependent influence.} Identical graph facts yield \emph{True} for \emph{Compose path} and \emph{False} for \emph{Read direct edge}: the path $A\to B\to C$ exists, whereas the direct edge $A\to C$ is absent. Under each instruction, source perturbations $\pm\delta$ produce responses at later sites that quantify influence, with intervening computation active. \textbf{C. Change the internal task state.} An early activation edit changes $h$ to $h'=h+\Delta_{\mathrm{task}}$ while the prompt and model weights remain fixed. The vector $\Delta_{\mathrm{task}}$ is the mean activation under \emph{Read fact 2} minus the mean activation under \emph{Read fact 1} on reference prompts. R1 and R2 denote two fixed attention heads. The schematic illustrates an answer change from \emph{True} (Fact 1) to \emph{False} (Fact 2), while the written instruction remains \emph{Read fact 1}.}
    \label{fig:overview}
\end{figure}

Here, we compare the same model components across tasks that use the same graph facts (Figure~\ref{fig:overview}A). For example, one task checks whether a path $A\to B\to C$ exists, while another checks whether the direct edge $A\to C$ exists (Figure~\ref{fig:overview}B). We measure each component's influence on later components and its importance for the answer. An early activation edit in Gemma-2-9B and Qwen3-14B switches the models between tasks that read different graph facts, while the task instruction and model parameters remain fixed (Figure~\ref{fig:overview}C). Our main findings are:
\begin{enumerate}
    \item \textbf{Influence changes with the task.} Across the five models, changing the task changes the influence pattern among the same selected components while the graph facts remain fixed.
    \item \textbf{Task context changes factual responses and influence.} In Gemma and Qwen, factual interventions affect the answer more strongly when the targeted graph fact is relevant to the task. An early activation edit also changes which graph facts the same heads respond to and how their outputs influence later components, while the task instruction stays fixed.
    \item \textbf{Task context improves prediction of later responses.} Within the same prompt, the response map for the edited or unedited condition predicts a later head's response to a separate factual intervention more accurately than the average of the two maps. In Gemma and Qwen fine-tuned on graph tasks, a fixed perturbation to an earlier head produces a larger mean squared response in a later head when the targeted graph fact can affect the answer.
\end{enumerate}

\section{Methods}

\subsection{Tasks and models}

Each prompt gives three graph facts: whether the directed edges $A\to B$, $B\to C$, and $A\to C$ are present or absent. The \emph{Compose path} task checks whether $A\to B$ and $B\to C$ are both present, forming the path $A\to B\to C$. The \emph{Read direct edge} task checks whether $A\to C$ is present. Both prompts in a pair use the same graph facts and the same response key, which maps \emph{True} and \emph{False} to answer tokens. Only the task instruction changes. We call graphs for which the two tasks require opposite answers opposite-answer graphs. This condition parallels incongruent trials in human decision-making, where different task rules favor opposite choices from the same stimuli (Figure~\ref{fig:overview}B) \citep{moneta2023task}.

The internal task edits and response prediction use two tasks. \emph{Read fact 1} checks whether $A\to B$ is present, and \emph{Read fact 2} checks whether $B\to C$ is present. The conditional-response analysis uses an \emph{AND} task: the answer is \emph{True} only when $B\to C$ and $A\to C$ are both present. Each edge can therefore affect the correct answer only when the other edge is present.

We measure influence in five open-weight models: Gemma-2-9B-it \citep{gemma2024gemma2}, Qwen3-14B \citep{yang2025qwen3}, Llama-3.1-8B-Instruct \citep{grattafiori2024llama3}, and the 1.5B-Deep-Instruct and 7B-Instruct variants of Falcon-H1 \citep{zuo2025falconh1}. Model selection during development considered graph-task performance and the identification of attention heads with measurable contributions to the answer. The five checkpoints span four model families and 1.5--14 billion parameters, allowing us to examine whether task-dependent influence recurs across families and sizes. We compare influence with component importance in Gemma, Qwen, Llama, and Falcon-H1-7B. Gemma and Qwen answer the fact-reading prompts correctly without additional training, so we use these models for factual interventions, internal task edits, and response prediction. The conditional-response analysis uses Gemma and Qwen fine-tuned on graph tasks. Model parameters remain fixed during all activation interventions. Within each analysis, we compare the same components across task conditions.

\subsection{Influence and importance}

We measure influence and importance using three attention heads at increasing depths, H1--H3, and one later residual-stream site, Res.\ (Figure~\ref{fig:overview}B). Both measurements use activations at the final input position, from which the model predicts the first answer token.

Influence is the signed response at a later site per unit source perturbation. Each site has a fixed task direction estimated from activation differences between the two instructions on reference prompts. Variation in the reference activations along this direction sets the scale: one standard deviation defines one unit. We apply separate positive and negative shifts of two units ($\delta=2$) to each source head's output along its task direction. All intervening computation remains active, so the response includes effects transmitted through other components. Importance is the decrease in the correct-answer logit margin when a site's full activation vector is replaced by its mean over reference prompts. The margin $m$ is the correct answer's logit minus the alternative answer's logit \citep{zhang2024patching}. For prompt $x$, the two effects are:
\[
I_{ij}(x)=
\frac{z_j^{+}(x;i)-z_j^{-}(x;i)}{2\delta},
\qquad
P_i(x)=m(x)-m_i^{\mathrm{replace}}(x).
\]
Here, $z_j^\pm(x;i)$ denotes the standardized activation projection at recipient $j$ after the positive or negative perturbation at source $i$. The term $m_i^{\mathrm{replace}}(x)$ denotes the margin after replacement at site $i$.

For each task, we average the effects over matched opposite-answer prompts. The six mean influence values and four mean replacement effects form the influence and importance profiles, respectively. For either profile, the angle between task means measures change in the relative pattern of effects (Appendix B.3). Each vocabulary group contains prompts that use the same names for the graph nodes. Angles are reported in degrees with 95\% confidence intervals from 10,000 bootstrap resamples of vocabulary groups.

\begin{figure}[t]
    \centering
    \includegraphics[width=\linewidth]{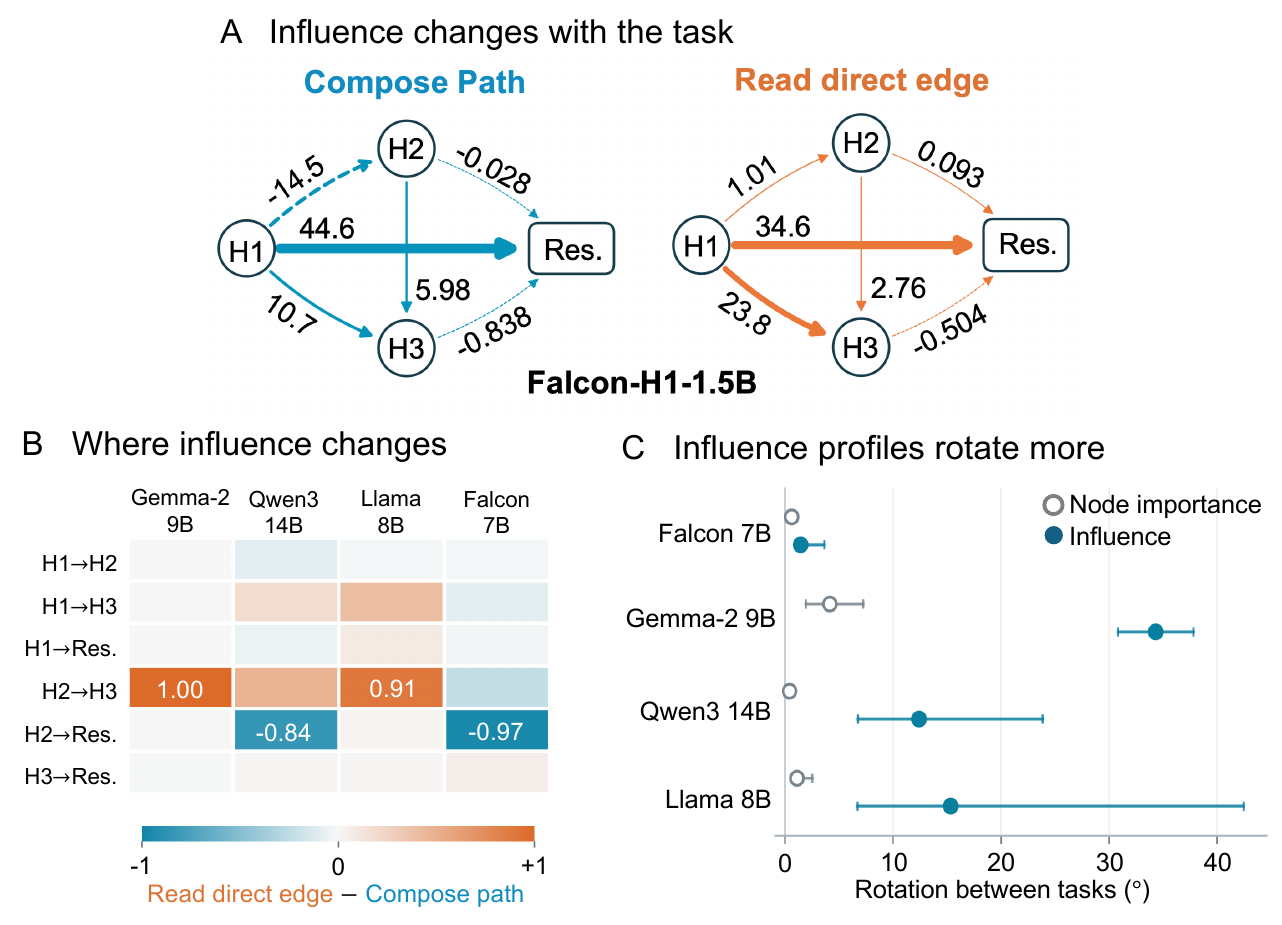}
    \caption{\textbf{Task changes redistribute influence among shared components.} \textbf{A. Influence changes with the task.} Falcon-H1-1.5B shows different influence patterns under \emph{Compose path} and \emph{Read direct edge} with identical graph facts. H1--H3 denote fixed attention heads in layer order; Res.\ denotes a later residual-stream site. Arrow labels give mean influence in units of $10^{-3}$, and widths indicate magnitude on a shared scale. Solid and dashed arrows denote positive and negative responses, respectively. \textbf{B. Where influence changes.} Color shows the difference in influence (\emph{Read direct edge} minus \emph{Compose path}) in four additional models. Each column is normalized to unit Euclidean norm, showing where the task-dependent change is concentrated. \textbf{C. Influence profiles rotate more.} Points show the angle between task profiles for the six influence values (filled) and four component-importance values (open). Whiskers show 95\% confidence intervals from 10,000 paired bootstrap resamples of vocabulary groups.}
    \label{fig:influence}
\end{figure}

\subsection{Factual interventions and downstream restoration}

Factual interventions change the residual activations associated with one graph fact. For example, reference runs differ only in whether ``$A\to B$ is present'' or ``$A\to B$ is absent.'' The intervention uses activation addition, which modifies a model's activations using differences between reference prompts \citep{turner2023activation}. We add the resulting activation differences to the residual states across the complete fact statement before a selected decoder layer. The same intervention is applied under both task instructions, while the original prompt remains fixed. Answer effects are changes in the logit difference between the labels representing \emph{True} and \emph{False}, with the sign reversed for present-to-absent interventions. Positive effects favor \emph{True} for an absent-to-present intervention and \emph{False} for a present-to-absent intervention. The two source controls each shuffle the activation difference's dimensions and randomly flip their signs, preserving its magnitude. The adjusted effect is the factual intervention's effect minus the mean source-control effect. A task is relevant when its correct answer depends on the targeted graph fact. Task selectivity is the adjusted effect under the relevant task minus that under the irrelevant task.

Downstream restoration uses activation replacement to test how later residual activations contribute to task selectivity \citep{vig2020mediation,zhang2024patching}. The factual intervention remains active while selected residual activations at a later layer are replaced with their values from the same prompt without the intervention. The two restoration controls each shuffle the dimensions and randomly flip the signs of the difference between the original and intervened downstream states, preserving its magnitude. We apply restoration and both restoration controls separately to runs with the factual intervention or either source control. The downstream contribution is the reduction in task selectivity under restoration minus the mean reduction under the restoration controls.

\subsection{Internal task edits and head contributions}

Internal task edits test how changes in task context affect factual responses and downstream influence while the written prompt remains fixed (Figure~\ref{fig:overview}C). The edit is a task-vector intervention, following work showing that internal task representations can induce task behavior \citep{hendel2023task,todd2024function}. The state $h$ contains residual activations before a selected decoder layer, covering the instruction position specifying which fact to read and all following input positions. The task vector $\Delta_{\mathrm{task}}$ is the mean reference activation for \emph{Read fact 2} minus that for \emph{Read fact 1}, following contrastive activation addition \citep{rimsky2024steering}. We add it to $h$ under \emph{Read fact 1} and subtract it under \emph{Read fact 2}. The reference graphs require the same answer under both instructions. The displacement remains fixed across changes to the graph facts.

\begin{figure}[!htbp]
    \centering
    \includegraphics[width=\linewidth]{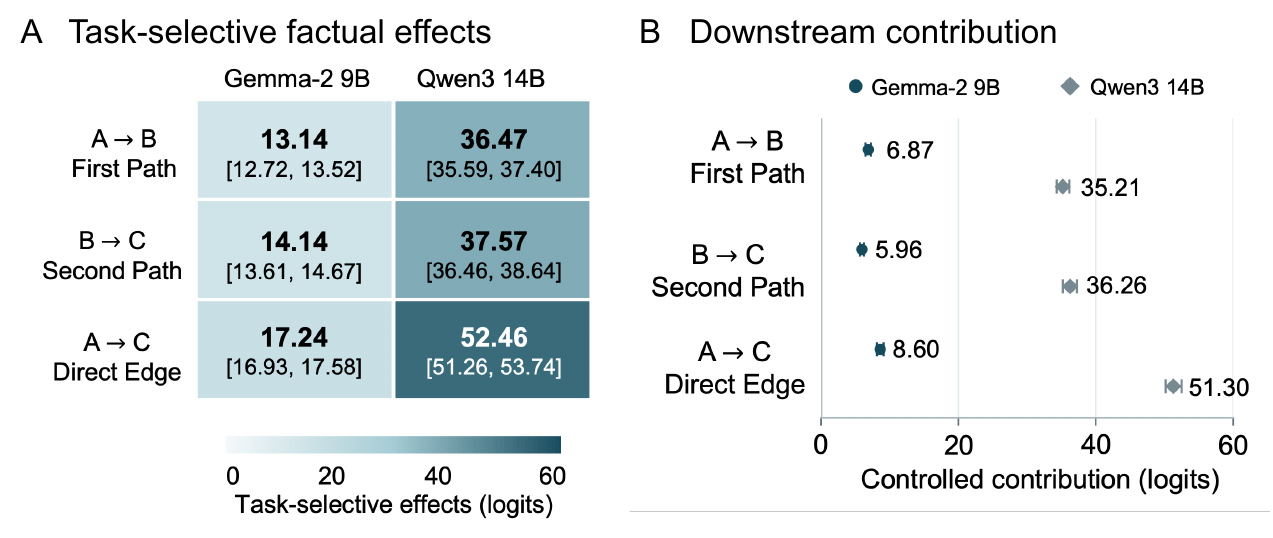}
    \caption{\textbf{Later activations support task-selective factual effects.} \textbf{A. Task-selective factual effects.} Cells show task selectivity in Gemma-2-9B and Qwen3-14B: the control-adjusted answer effect under the relevant instruction minus that under the irrelevant instruction. The edges $A\to B$ and $B\to C$ are relevant to \emph{Compose path}, whereas $A\to C$ is relevant to \emph{Read direct edge}. \textbf{B. Downstream contribution.} The factual intervention remains active while selected later residual activations are restored to their values from the same prompt without the intervention. Points show the reduction in task selectivity under restoration minus the mean reduction under restoration controls. Brackets in A and whiskers in B show pointwise 95\% confidence intervals from 10,000 paired bootstrap resamples of vocabulary groups.}
    \label{fig:factual_sources}
\end{figure}

For each vocabulary group, $E_f$ is the mean squared distance between a head's outputs for paired prompts with fact $f$ absent and present. The instruction and other facts are fixed within each pair, and outputs are measured at the final input position. Each fact's response share is $E_f/\sum_{g=1}^{3}E_g$, with all three facts included in the denominator. An analysis of all attention heads measures how the edit changes the balance between responses to the alternative and originally requested facts:
\[
Q=
\frac{
(E_{\mathrm{alt}}^{\mathrm{edit}}-E_{\mathrm{orig}}^{\mathrm{edit}})
-
(E_{\mathrm{alt}}^{\mathrm{unedited}}-E_{\mathrm{orig}}^{\mathrm{unedited}})
}{
\sum_{f=1}^{3}
(E_f^{\mathrm{unedited}}+E_f^{\mathrm{edit}})
}.
\]

The influence analysis follows the same four selected sites before and after the task edit. The fixed factual directions used to perturb source heads and measure recipient responses are estimated from present-minus-absent reference activations under each fact-reading instruction. The responses for the two factual directions are averaged to form a six-pair influence profile. Its target is the unedited profile under the alternative instruction with the same graph facts. The sum of squared differences across the six influence values measures distance to this target. We compare distances before the edit, under the task edit, and under two random early edits.

\begin{figure}[!b]
    \centering
    \includegraphics[width=\linewidth]{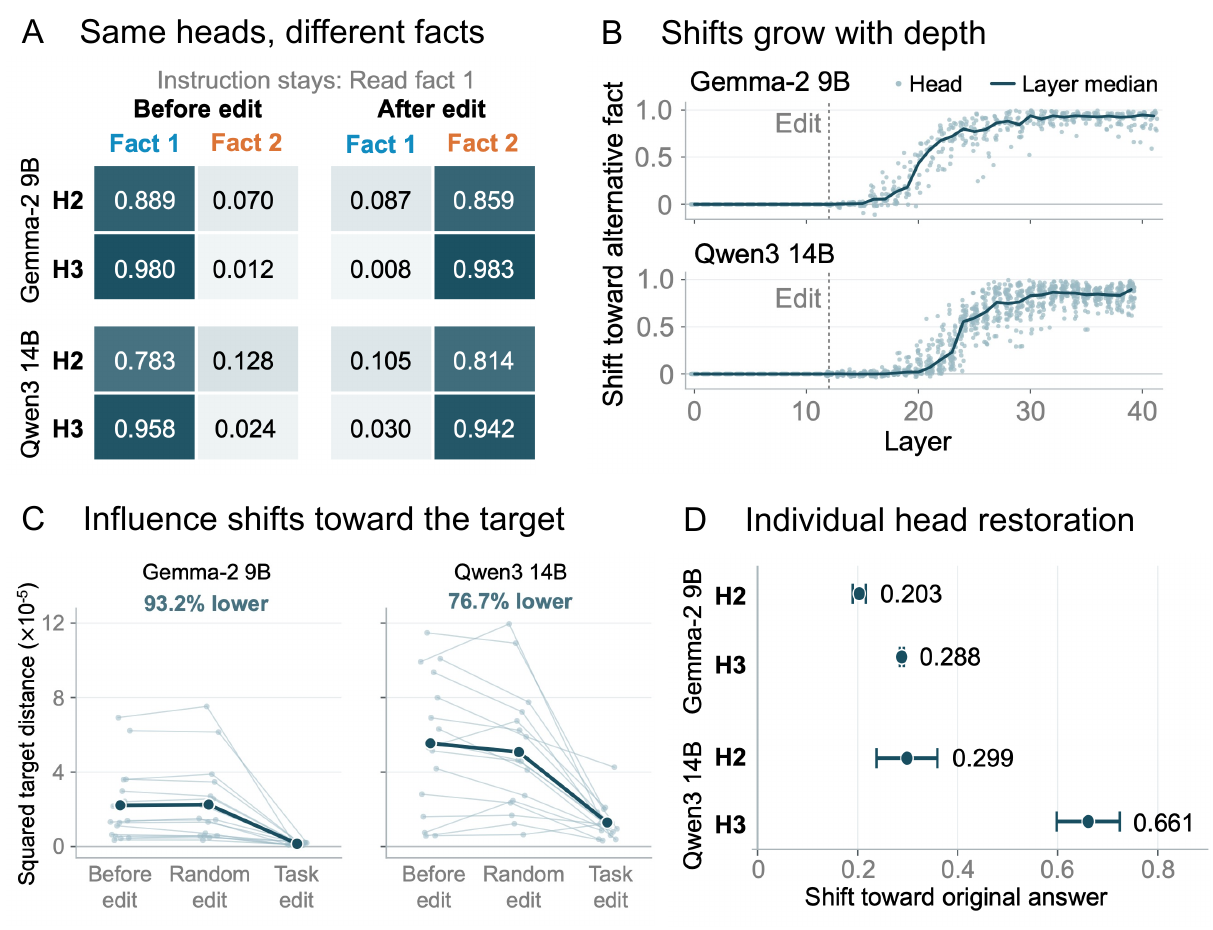}
    \caption{\textbf{Internal task edits redirect factual responses and influence.} Released Gemma-2-9B and Qwen3-14B weights remain fixed during all interventions. \textbf{A. Same heads, different facts.} Cells show mean response shares at fixed heads H2 and H3 before and after the early edit, with the written instruction held at \emph{Read fact 1}. Shares are normalized over squared head-output responses to changes in all three facts. \textbf{B. Shifts grow with depth.} Dots show each head's mean factual-reassignment score, and lines show layer medians across all 672 Gemma heads and 1,600 Qwen heads. Dashed lines mark the edit layer. \textbf{C. Influence shifts toward the target.} Squared distances compare the six-pair influence profile with the unedited profile under the alternative instruction. Thin lines represent vocabulary groups, and thick lines show means. \textbf{D. Individual head restoration.} Points show control-adjusted increases in the original-answer margin, in logits, after individual H2 or H3 restoration while the early edit remains active. Whiskers show approximate simultaneous 95\% confidence intervals.}
    \label{fig:task_edit}
\end{figure}

Individual head restoration tests how H2 and H3 contribute to the answer after the task edit. Each head is restored to its output from the same prompt without the edit, while the early edit remains active. On graphs where the two instructions require opposite answers, the restoration effect is the increase in the logit margin for the answer required by the written instruction. The adjusted restoration effect is the restoration effect minus the mean effect of two random changes to the head's output. The controlled contribution is the adjusted effect under the task edit minus the mean adjusted effect under the two random early edits.

\subsection{Prediction and conditional responses}

In the tested Gemma and Qwen models, we use responses to controlled H2 perturbations to predict how a factual intervention at the same head affects the downstream head H3. For each prompt and factual direction $k$, we subtract the output under the negative H2 perturbation from the output under the positive perturbation. The resulting differences are $\mathbf a_k$ at H2 and $\mathbf b_k$ at H3. The response map averages the two directional maps:
\[
M=\frac{1}{2}\sum_{k=1}^{2}
\frac{\mathbf b_k\mathbf a_k^\top}{\|\mathbf a_k\|_2^2}.
\]
The unedited and task-edited runs provide one map each for the same prompt. A separate factual intervention replaces H2's output with its value from a reference run in which one graph fact has changed. We compare predictions from the map for the current internal context and the average of the two maps, using the same measured H2 displacement. Prediction error is the squared distance between the predicted and observed H3 output changes.

The conditional-response analysis changes factual relevance under AND while keeping the H2 displacement fixed in the fine-tuned Gemma and Qwen models. We derive the displacement from H2's output difference when the targeted edge changes from absent to present while the other edge is absent. The same displacement is added at H2 with the other edge either present or absent. The targeted edge is initially absent in both contexts, so both prompts require \emph{False} before intervention. Response strength is the squared change in H3's output relative to its unperturbed value in each context. One control permutes the displacement's dimensions and the other randomly flips their signs, preserving the magnitude in both cases. The controlled effect is the relevant-minus-irrelevant response difference minus the mean corresponding difference under the permutation and sign-flip controls.

\section{Results}

\subsection{Task changes redistribute influence}

A perturbation of the same head produces different downstream responses under the two task instructions, with the graph facts held fixed. In Falcon-H1-1.5B, H1's mean influence on H2 changes from $-0.0145$ under \emph{Compose path} to $0.00101$ under \emph{Read direct edge} (Figure~\ref{fig:influence}A). The same instruction change more than doubles H1's influence on H3 ($0.0107$ to $0.0238$), while reducing its influence on the later residual site ($0.0446$ to $0.0346$).

The largest influence changes in the four additional models involve H2 (Figure~\ref{fig:influence}B). H2’s influence on H3 increases from \emph{Compose path} to \emph{Read direct edge} in Gemma-2-9B and Llama-3.1-8B. The same instruction change reduces H2’s influence on the later residual site in Qwen3-14B and Falcon-H1-7B.

Component importance provides a second view of the task change: how much each selected site contributes to the answer. With the residual site included, the six-entry influence profiles rotate more between tasks than the four-site importance profiles in all four models (Figure~\ref{fig:influence}C). The paired influence-minus-importance angle differences are $30.17^\circ\;[25.54^\circ,\,34.81^\circ]$ in Gemma and $11.98^\circ\;[5.35^\circ,\,26.42^\circ]$ in Qwen. The differences in Llama and Falcon-H1-7B are $14.22^\circ\;[3.53^\circ,\,48.22^\circ]$ and $0.83^\circ\;[0.07^\circ,\,3.51^\circ]$, respectively. All four Bonferroni-adjusted bootstrap intervals exclude zero, with approximate simultaneous 95\% coverage. The residual site contributes 97.8--99.95\% of the importance profiles' squared norm. In an exploratory comparison restricted to the three heads, influence angles are smaller than importance angles in Gemma and Falcon-H1-7B and larger in Qwen and Llama. A small importance angle does not imply little change in absolute answer effects.

\begin{figure}[!b]
    \centering
    \includegraphics[width=\linewidth]{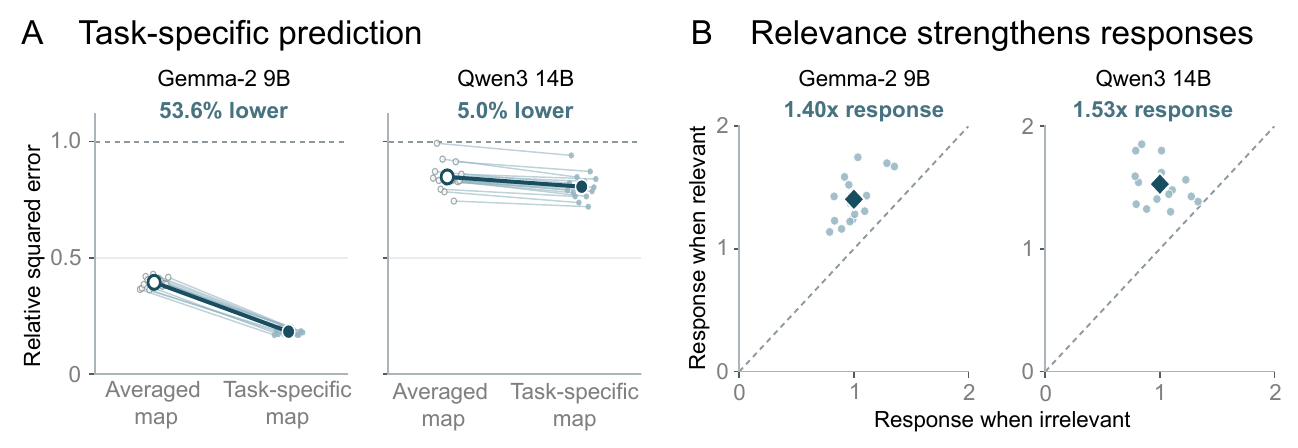}
    \caption{\textbf{Context shapes prediction and downstream responses.} \textbf{A. Task-specific prediction.} Response maps in released Gemma-2-9B and Qwen3-14B predict the H3 output change caused by a separate factual intervention at H2. Errors are divided by each model's mean error for predicting no H3 change, shown by the dashed line at 1. Thin lines join paired vocabulary means, and thick lines show overall means. Percentages give reductions in mean squared error relative to the averaged map. \textbf{B. Relevance strengthens responses.} Gemma and Qwen fine-tuned on graph tasks receive the same H2 displacement in the relevant and irrelevant contexts under AND. Both axes show squared H3 responses divided by the model's mean response in the irrelevant context. Circles represent vocabulary means, diamonds show overall means, and the dashed diagonal marks equal responses.}
    \label{fig:prediction_conditional}
\end{figure}

\subsection{Later activations support task-selective factual effects}

Factual interventions have stronger control-adjusted answer effects when the targeted fact determines the required answer (Figure~\ref{fig:factual_sources}A). For $A\to B$, $B\to C$, and $A\to C$, task selectivity in Gemma-2-9B is $13.14\;[12.52,\,13.68]$, $14.14\;[13.39,\,14.90]$, and $17.24\;[16.81,\,17.73]$ logits, respectively. The corresponding effects in Qwen3-14B are $36.47\;[35.24,\,37.86]$, $37.57\;[35.93,\,39.04]$, and $52.46\;[50.83,\,54.36]$ logits.

Downstream restoration reduces task selectivity more than the restoration controls for all three facts in both models (Figure~\ref{fig:factual_sources}B). In the same edge order, the control-adjusted reductions are $6.87\;[6.28,\,7.39]$, $5.96\;[5.52,\,6.37]$, and $8.60\;[7.97,\,9.33]$ logits in Gemma. The reductions in Qwen are $35.21\;[33.97,\,36.59]$, $36.26\;[34.68,\,37.64]$, and $51.30\;[49.74,\,53.08]$ logits. Bonferroni-corrected bootstrap intervals exclude zero for all selectivity and restoration effects, with approximate simultaneous 95\% coverage.

\subsection{Internal task edits redirect factual responses and influence}

An early activation edit redirects factual responses in Gemma-2-9B and Qwen3-14B. Our analysis uses the induced task switch to measure changes in fixed components' factual responses and downstream influence. Across both edit directions, each model switches to the alternative instruction's answer on all 512 opposite-answer prompts and preserves the correct answer on all 512 same-answer prompts. Both random early edits preserve the original answers on all 1,024 prompts per model. Under the written \emph{Read fact 1} instruction, H2 and H3 become most responsive to Fact 2 (Figure~\ref{fig:task_edit}A). Fact 2's response share increases by $78.9\;[78.5,\,79.4]$ percentage points at H2 and $97.1\;[96.8,\,97.4]$ at H3 in Gemma. Qwen's corresponding increases are $68.7\;[67.5,\,69.7]$ and $91.8\;[91.1,\,92.6]$ percentage points. All four corrected bootstrap intervals exclude zero, with approximate simultaneous 95\% coverage. Across all attention heads, both models meet the criterion of positive median reassignment in all three prespecified post-edit depth bands, including after adjustment for random early edits (Figure~\ref{fig:task_edit}B). Late-band medians are $0.933\;[0.930,\,0.935]$ in Gemma and $0.850\;[0.847,\,0.854]$ in Qwen, with pointwise 97.5\% bootstrap intervals.

Mean squared distance to the unedited influence profile under the alternative instruction falls by 93.2\% in Gemma and 76.7\% in Qwen after the edit (Figure~\ref{fig:task_edit}C). Mean distances are also lower than under the random early edits in both models. Individual head restoration produces small control-adjusted increases in the original-answer margin while the early edit remains active (Figure~\ref{fig:task_edit}D). The H2 and H3 effects are $0.203\;[0.190,\,0.216]$ and $0.288\;[0.284,\,0.292]$ logits in Gemma, respectively. Qwen's corresponding effects are $0.299\;[0.238,\,0.359]$ and $0.661\;[0.598,\,0.724]$ logits. All four approximate simultaneous 95\% intervals exclude zero. Relative to the mean edit-induced margin shift, the restoration effects are $1.13\%\;[1.08\%,\,1.18\%]$ and $1.61\%\;[1.59\%,\,1.62\%]$ in Gemma, and $0.62\%\;[0.54\%,\,0.70\%]$ and $1.37\%\;[1.29\%,\,1.46\%]$ in Qwen.

\subsection{Context shapes prediction and downstream responses}

Within the same prompt, the map for the current internal context predicts factual responses more accurately than the average of the unedited and task-edited maps (Figure~\ref{fig:prediction_conditional}A). Mean squared error falls by 53.6\% in Gemma-2-9B and 5.0\% in Qwen3-14B. In fine-tuned Gemma and Qwen, the same H2 displacement produces stronger H3 responses when the targeted fact is relevant under AND (Figure~\ref{fig:prediction_conditional}B). The relevant-to-irrelevant ratios of mean squared H3 responses are $1.40\;[1.31,\,1.51]$ in Gemma and $1.53\;[1.34,\,1.74]$ in Qwen. Both corrected bootstrap intervals exclude 1, with approximate simultaneous 95\% coverage.

\section{Discussion}

Here we have shown that shared components can change both which facts they respond to and how they influence later computation. Influence and answer importance provide complementary descriptions of component roles, as their relative patterns change by different amounts across tasks. In Gemma and Qwen, an early activation edit shifts factual responses and influence while the written instruction remains fixed. The measured roles therefore depend on the model's internal task state. This dependence also matters for prediction: response maps matched to the current state predict downstream factual effects more accurately within the same prompt.

\textbf{Network interpretation.} Our findings suggest that a shared component's role depends on the state of the network receiving its output. Interventions on reusable inhibition circuits can change which tokens later heads select, providing a mechanistic account of how one component shapes another's computation \citep{merullo2024circuit}. Communication between layers can also rely on a few directions in the residual stream that carry features read by later layers \citep{merullo2024talking}. In fine-tuned Gemma and Qwen, the surrounding facts change the downstream response to an identical head displacement. The mean squared response is larger when the targeted fact determines the AND answer. A related principle appears in recurrent models of context-dependent decisions, where network dynamics determine how the same sensory input contributes to a choice \citep{mante2013context}. Both settings illustrate how context shapes information processing within a shared network. In the brain, a region's task-related activity can be predicted from activity in other regions weighted by resting-state functional connectivity \citep{cole2016activity}. In Gemma and Qwen without additional training, task-specific maps improve prediction of downstream responses within the same prompt (Figure~\ref{fig:prediction_conditional}A). Task context therefore helps explain how an intervention propagates, even when the intervened component and its imposed change remain fixed.

\textbf{Applicability and scope.} The graph setting makes factual relevance explicit, allowing comparisons that keep the available information fixed while changing how it should be used. Here, each influence value describes a response transmitted through the intervening network, including components outside the selected sites. Its interpretation is tied to the chosen activation direction and perturbation size. Our prediction analysis uses the measured source displacement and a response map estimated for the same prompt, so the improvement shows the benefit of matching the internal task state within that setting. The conditional-response finding applies to Gemma and Qwen fine-tuned on graph tasks and evaluated under the AND instruction. Factual interventions, internal task edits, and response prediction use Gemma and Qwen without additional training.

\textbf{Future directions.} A central question is whether influence measurements can predict responses beyond the prompts on which they were estimated. Response maps could be evaluated on new graphs and instructions without recalibration for each prompt, testing how much of the measured interaction is shared across contexts. A finer description of the components may help explain which interactions transfer. Sparse feature circuits trace causal interactions among interpretable activation features \citep{marks2025sparse}. Sparse approximations of feedforward layers also allow interactions between features to be traced through those layers \citep{dunefsky2024transcoders}. The comparison of features across tasks could reveal whether a shared head uses different features or whether shared features have different effects on later computation. The same distinction could be examined in longer passages, where factual relevance depends on several statements and intermediate reasoning steps. These extensions would test whether context-dependent interactions can predict which information contributes to an answer as task demands change.

\section*{Acknowledgments}

S.M. and D.S.B. acknowledge support from the Army Research Office MURI program (through grant number W911NF2410228).

\section*{AI Use Disclosure}

In this work, we used generative AI tools for initial literature review, code assistance and updating parts of the manuscript to have better readability. We have reviewed all AI-assisted works including the initial literature review, and the code was verified and tested for correctness. We take responsibility for the final content of this work, including text, claims or artifacts produced with the aid of generative AI.

\section*{Reproducibility Statement}
The experimental setup and measurements are described in Section 2, with graph construction, prompt examples, and evaluation sets in Appendix A. Appendices B–F provide the component-selection procedures, intervention and control definitions, response-prediction methods, and fine-tuning settings. Appendix G describes the aggregation procedures and statistical uncertainty estimates. The code for all the experiments and analyses in the paper will be released following the review process.

\bibliographystyle{iclr2027_conference}
\bibliography{references}

\begin{thebibliography}{30}
\providecommand{\natexlab}[1]{#1}
\providecommand{\url}[1]{\texttt{#1}}
\expandafter\ifx\csname urlstyle\endcsname\relax
  \providecommand{\doi}[1]{doi: #1}\else
  \providecommand{\doi}{doi: \begingroup \urlstyle{rm}\Url}\fi

\bibitem[Bassett \& Sporns(2017)Bassett and Sporns]{bassett2017network}
Danielle~S. Bassett and Olaf Sporns.
\newblock Network neuroscience.
\newblock \emph{Nature Neuroscience}, 20\penalty0 (3):\penalty0 353--364, 2017.

\bibitem[Brown et~al.(2020)Brown, Mann, Ryder, et~al.]{brown2020fewshot}
Tom Brown, Benjamin Mann, Nick Ryder, et~al.
\newblock Language models are few-shot learners.
\newblock In H.~Larochelle, M.~Ranzato, R.~Hadsell, M.~F. Balcan, and H.~Lin
  (eds.), \emph{Advances in Neural Information Processing Systems}, volume~33,
  pp.\  1877--1901. Curran Associates, Inc., 2020.

\bibitem[Cho et~al.(2026)Cho, Yang, Minegishi, and Inoue]{cho2026removal}
Hakaze Cho, Haolin Yang, Gouki Minegishi, and Naoya Inoue.
\newblock Mechanism of task-oriented information removal in in-context
  learning.
\newblock In \emph{International Conference on Learning Representations}, pp.\
  9512--9598, 2026.

\bibitem[Cole et~al.(2013)Cole, Reynolds, Power, Repovs, Anticevic, and
  Braver]{cole2013multitask}
Michael~W. Cole, Jeremy~R. Reynolds, Jonathan~D. Power, Grega Repovs, Alan
  Anticevic, and Todd~S. Braver.
\newblock Multi-task connectivity reveals flexible hubs for adaptive task
  control.
\newblock \emph{Nature Neuroscience}, 16\penalty0 (9):\penalty0 1348--1355,
  2013.

\bibitem[Cole et~al.(2016)Cole, Ito, Bassett, and Schultz]{cole2016activity}
Michael~W. Cole, Takuya Ito, Danielle~S. Bassett, and Douglas~H. Schultz.
\newblock Activity flow over resting-state networks shapes cognitive task
  activations.
\newblock \emph{Nature Neuroscience}, 19\penalty0 (12):\penalty0 1718--1726,
  2016.
\newblock \doi{10.1038/nn.4406}.

\bibitem[Conmy et~al.(2023)Conmy, Mavor-Parker, Lynch, Heimersheim, and
  Garriga-Alonso]{conmy2023automated}
Arthur Conmy, Augustine Mavor-Parker, Aengus Lynch, Stefan Heimersheim, and
  Adri\`{a} Garriga-Alonso.
\newblock Towards automated circuit discovery for mechanistic interpretability.
\newblock In A.~Oh, T.~Naumann, A.~Globerson, K.~Saenko, M.~Hardt, and
  S.~Levine (eds.), \emph{Advances in Neural Information Processing Systems},
  volume~36, pp.\  16318--16352. Curran Associates, Inc., 2023.

\bibitem[Dunefsky et~al.(2024)Dunefsky, Chlenski, and
  Nanda]{dunefsky2024transcoders}
Jacob Dunefsky, Philippe Chlenski, and Neel Nanda.
\newblock Transcoders find interpretable {LLM} feature circuits.
\newblock In \emph{Advances in Neural Information Processing Systems},
  volume~37, pp.\  24375--24410, 2024.
\newblock \doi{10.52202/079017-0768}.

\bibitem[{Gemma Team}(2024)]{gemma2024gemma2}
{Gemma Team}.
\newblock {Gemma 2}: Improving open language models at a practical size.
\newblock arXiv preprint arXiv:2408.00118, 2024.

\bibitem[Gonzalez-Castillo et~al.(2015)Gonzalez-Castillo, Hoy, Handwerker,
  Robinson, Buchanan, Saad, and Bandettini]{gonzalezcastillo2015tracking}
Javier Gonzalez-Castillo, Colin~W. Hoy, Daniel~A. Handwerker, Meghan~E.
  Robinson, Laura~C. Buchanan, Ziad~S. Saad, and Peter~A. Bandettini.
\newblock Tracking ongoing cognition in individuals using brief, whole-brain
  functional connectivity patterns.
\newblock \emph{Proceedings of the National Academy of Sciences}, 112\penalty0
  (28):\penalty0 8762--8767, 2015.

\bibitem[Grattafiori et~al.(2024)Grattafiori, Dubey, Jauhri,
  et~al.]{grattafiori2024llama3}
Aaron Grattafiori, Abhimanyu Dubey, Abhinav Jauhri, et~al.
\newblock The {Llama 3} herd of models.
\newblock arXiv preprint arXiv:2407.21783, 2024.

\bibitem[Han et~al.(2026)Han, Andreas, Fedorenko, and de~Varda]{han2026modular}
Pengrui Han, Jacob Andreas, Evelina Fedorenko, and Andrea~Gregor de~Varda.
\newblock Modular cognitive architecture emerges in large language models.
\newblock arXiv preprint arXiv:2608.13567, 2026.

\bibitem[Hendel et~al.(2023)Hendel, Geva, and Globerson]{hendel2023task}
Roee Hendel, Mor Geva, and Amir Globerson.
\newblock In-context learning creates task vectors.
\newblock In \emph{Findings of the Association for Computational Linguistics:
  EMNLP 2023}, pp.\  9318--9333. Association for Computational Linguistics,
  2023.

\bibitem[Hu et~al.(2022)Hu, Shen, Wallis, Allen-Zhu, Li, Wang, Wang, and
  Chen]{hu2022lora}
Edward~J. Hu, Yelong Shen, Phillip Wallis, Zeyuan Allen-Zhu, Yuanzhi Li, Shean
  Wang, Lu~Wang, and Weizhu Chen.
\newblock Lo{RA}: Low-rank adaptation of large language models.
\newblock In \emph{International Conference on Learning Representations}, 2022.

\bibitem[Krienen et~al.(2014)Krienen, Yeo, and
  Buckner]{krienen2014reconfigurable}
Fenna~M. Krienen, B.~T.~Thomas Yeo, and Randy~L. Buckner.
\newblock Reconfigurable task-dependent functional coupling modes cluster
  around a core functional architecture.
\newblock \emph{Philosophical Transactions of the Royal Society B: Biological
  Sciences}, 369\penalty0 (1653):\penalty0 20130526, 2014.

\bibitem[Liu et~al.(2025)Liu, Liu, Wu, Ning, Sun, Xia, Yang, Gao, Qiang, Ge,
  Liu, Han, and Hu]{liu2025brainfunctional}
Yiheng Liu, Zhengliang Liu, Zihao Wu, Junhao Ning, Haiyang Sun, Sichen Xia,
  Yang Yang, Xiaohui Gao, Ning Qiang, Bao Ge, Tianming Liu, Junwei Han, and
  Xintao Hu.
\newblock Brain-inspired exploration of functional networks and key neurons in
  large language models.
\newblock arXiv preprint arXiv:2502.20408, 2025.

\bibitem[Mante et~al.(2013)Mante, Sussillo, Shenoy, and
  Newsome]{mante2013context}
Valerio Mante, David Sussillo, Krishna~V. Shenoy, and William~T. Newsome.
\newblock Context-dependent computation by recurrent dynamics in prefrontal
  cortex.
\newblock \emph{Nature}, 503\penalty0 (7474):\penalty0 78--84, 2013.

\bibitem[Marks et~al.(2025)Marks, Rager, Michaud, Belinkov, Bau, and
  Mueller]{marks2025sparse}
Samuel Marks, Can Rager, Eric~J. Michaud, Yonatan Belinkov, David Bau, and
  Aaron Mueller.
\newblock Sparse feature circuits: Discovering and editing interpretable causal
  graphs in language models.
\newblock In \emph{International Conference on Learning Representations}, pp.\
  23888--23923, 2025.

\bibitem[Merullo et~al.(2024{\natexlab{a}})Merullo, Eickhoff, and
  Pavlick]{merullo2024circuit}
Jack Merullo, Carsten Eickhoff, and Ellie Pavlick.
\newblock Circuit component reuse across tasks in transformer language models.
\newblock In \emph{International Conference on Learning Representations}, pp.\
  18349--18377, 2024{\natexlab{a}}.

\bibitem[Merullo et~al.(2024{\natexlab{b}})Merullo, Eickhoff, and
  Pavlick]{merullo2024talking}
Jack Merullo, Carsten Eickhoff, and Ellie Pavlick.
\newblock Talking heads: Understanding inter-layer communication in transformer
  language models.
\newblock In A.~Globerson, L.~Mackey, D.~Belgrave, A.~Fan, U.~Paquet,
  J.~Tomczak, and C.~Zhang (eds.), \emph{Advances in Neural Information
  Processing Systems}, volume~37, pp.\  61372--61418. Curran Associates, Inc.,
  2024{\natexlab{b}}.

\bibitem[Moneta et~al.(2023)Moneta, Garvert, Heekeren, and
  Schuck]{moneta2023task}
Nir Moneta, Mona~M. Garvert, Hauke~R. Heekeren, and Nicolas~W. Schuck.
\newblock Task state representations in {vmPFC} mediate relevant and irrelevant
  value signals and their behavioral influence.
\newblock \emph{Nature Communications}, 14:\penalty0 3156, 2023.
\newblock \doi{10.1038/s41467-023-38709-w}.

\bibitem[Rimsky et~al.(2024)Rimsky, Gabrieli, Schulz, Tong, Hubinger, and
  Turner]{rimsky2024steering}
Nina Rimsky, Nick Gabrieli, Julian Schulz, Meg Tong, Evan Hubinger, and
  Alexander~Matt Turner.
\newblock Steering {Llama 2} via contrastive activation addition.
\newblock In \emph{Proceedings of the 62nd Annual Meeting of the Association
  for Computational Linguistics (Volume 1: Long Papers)}, pp.\  15504--15522.
  Association for Computational Linguistics, 2024.

\bibitem[Todd et~al.(2024)Todd, Li, Sen~Sharma, Mueller, Wallace, and
  Bau]{todd2024function}
Eric Todd, Millicent~L. Li, Arnab Sen~Sharma, Aaron Mueller, Byron~C. Wallace,
  and David Bau.
\newblock Function vectors in large language models.
\newblock In \emph{International Conference on Learning Representations}, pp.\
  17282--17333, 2024.

\bibitem[Turner et~al.(2023)Turner, Thiergart, Udell, Leech, Mini, and
  MacDiarmid]{turner2023activation}
Alex Turner, Lisa Thiergart, David Udell, Gavin Leech, Ulisse Mini, and Monte
  MacDiarmid.
\newblock Activation addition: Steering language models without optimization.
\newblock arXiv preprint arXiv:2308.10248v1, 2023.

\bibitem[Vig et~al.(2020)Vig, Gehrmann, Belinkov, Qian, Nevo, Singer, and
  Shieber]{vig2020mediation}
Jesse Vig, Sebastian Gehrmann, Yonatan Belinkov, Sharon Qian, Daniel Nevo,
  Yaron Singer, and Stuart Shieber.
\newblock Investigating gender bias in language models using causal mediation
  analysis.
\newblock In H.~Larochelle, M.~Ranzato, R.~Hadsell, M.~F. Balcan, and H.~Lin
  (eds.), \emph{Advances in Neural Information Processing Systems}, volume~33,
  pp.\  12388--12401. Curran Associates, Inc., 2020.

\bibitem[Wang et~al.(2023)Wang, Variengien, Conmy, Shlegeris, and
  Steinhardt]{wang2023ioi}
Kevin Wang, Alexandre Variengien, Arthur Conmy, Buck Shlegeris, and Jacob
  Steinhardt.
\newblock Interpretability in the wild: A circuit for indirect object
  identification in {GPT-2} small.
\newblock In \emph{International Conference on Learning Representations}, 2023.

\bibitem[Wei et~al.(2022)Wei, Bosma, Zhao, Guu, Yu, Lester, Du, Dai, and
  Le]{wei2022finetuned}
Jason Wei, Maarten Bosma, Vincent~Y. Zhao, Kelvin Guu, Adams~Wei Yu, Brian
  Lester, Nan Du, Andrew~M. Dai, and Quoc~V. Le.
\newblock Finetuned language models are zero-shot learners.
\newblock In \emph{International Conference on Learning Representations}, 2022.

\bibitem[Yang et~al.(2025)Yang, Li, Yang, et~al.]{yang2025qwen3}
An~Yang, Anfeng Li, Baosong Yang, et~al.
\newblock {Qwen3} technical report.
\newblock arXiv preprint arXiv:2505.09388, 2025.

\bibitem[Yang et~al.(2026)Yang, Cho, and Inoue]{yang2026localizing}
Haolin Yang, Hakaze Cho, and Naoya Inoue.
\newblock Localizing task recognition and task learning in in-context learning
  via attention head analysis.
\newblock In \emph{International Conference on Learning Representations}, pp.\
  93780--93839, 2026.

\bibitem[Zhang \& Nanda(2024)Zhang and Nanda]{zhang2024patching}
Fred Zhang and Neel Nanda.
\newblock Towards best practices of activation patching in language models:
  Metrics and methods.
\newblock In \emph{International Conference on Learning Representations}, pp.\
  1651--1678, 2024.

\bibitem[Zuo et~al.(2025)Zuo, Velikanov, Chahed, et~al.]{zuo2025falconh1}
Jingwei Zuo, Maksim Velikanov, Ilyas Chahed, et~al.
\newblock {Falcon-H1}: A family of hybrid-head language models redefining
  efficiency and performance.
\newblock arXiv preprint arXiv:2507.22448, 2025.

\end{thebibliography}

\clearpage
\appendix
\setcounter{figure}{0}
\renewcommand{\thefigure}{S\arabic{figure}}
\renewcommand{\theHfigure}{supp.\arabic{figure}}
\setcounter{table}{0}
\renewcommand{\thetable}{S\arabic{table}}
\renewcommand{\theHtable}{supp.\arabic{table}}
\section*{Appendix: Supplementary Methods and Results}

\section{Tasks, prompts, and evaluation sets}

\subsection{Task construction and prompt examples}

Each prompt contains a graph description, a task instruction, and an answer rule. The symbols $A,B,C$ stand for node names, with each vocabulary group using one set of names. The three edges can each be present or absent, giving eight possible graphs. The instruction determines the required \emph{True} or \emph{False} answer. The examples below use a response key that maps the answer to \texttt{A} or \texttt{B}, with a reversed key exchanging the labels. The influence--importance analysis instead uses \texttt{Yes}/\texttt{No} labels (Appendix~A.2). Paired task prompts preserve the graph description and response key, so only the requested task changes. The factual-intervention analysis also uses two fact wordings: \textquotedblleft{}$A\to B$: absent\textquotedblright{} and \textquotedblleft{}There is no edge from $A$ to $B$\textquotedblright{}. The graph facts, task question, and response key remain the same across the two wordings.

For \emph{Compose path} and \emph{Read direct edge}, four graphs require opposite answers and four require the same answer. The opposite-answer graphs have a complete path without the direct edge, or the direct edge without a complete path. For the two fact-reading instructions, the answers differ when one requested edge is present and the other is absent. The remaining graphs have both requested edges present or both absent. The following \emph{Compose path} example requires \texttt{A} (\emph{True}), since both path edges are present.

\noindent\textbf{System message}
\begin{quote}
\ttfamily\raggedright
\noindent\detokenize{Evaluate the stated facts literally. Use only the explicit graph facts and the requested operation.}\par
\end{quote}

\noindent\textbf{User message}
\begin{quote}
\ttfamily\raggedright
\noindent\detokenize{The directed graph has the following explicit edge facts.}\par
\noindent\detokenize{Pibepaka -> Kupepezi: present.}\par
\noindent\detokenize{Pibepaka -> Karaduki: absent.}\par
\noindent\detokenize{Kupepezi -> Karaduki: present.}\par
\noindent\detokenize{Are both Pibepaka -> Kupepezi and Kupepezi -> Karaduki present?}\par
\noindent\detokenize{Response key: A means true; B means false. Reply with exactly A or B. Do not explain.}\par
\end{quote}

The paired \emph{Read direct edge} prompt replaces only the question with \textquotedblleft{}Is the explicit edge \texttt{Pibepaka -> Karaduki} present?\textquotedblright{} The correct response becomes \texttt{B} (\emph{False}), since the direct edge is absent.

The internal-edit and response-prediction prompts number the graph facts and place the response key before the question. The following \emph{Read fact 1} example uses the same system message. The paired \emph{Read fact 2} prompt changes only \texttt{1} to \texttt{2} in the question, changing the correct response from \texttt{A} to \texttt{B}. The fact numbers remain attached to the same edges when the statements are reordered.

\noindent\textbf{User message}
\begin{quote}
\ttfamily\raggedright
\noindent\detokenize{Each numbered fact gives the presence or absence of one directed edge.}\par
\noindent\detokenize{Fact 1: Perepoda -> Botageto is present.}\par
\noindent\detokenize{Fact 2: Botageto -> Kubeluso is absent.}\par
\noindent\detokenize{Fact 3: Perepoda -> Kubeluso is absent.}\par
\noindent\detokenize{Response key: A means true; B means false. Reply with exactly A or B. Do not explain.}\par
\noindent\detokenize{Question: Is the edge in Fact 1 present?}\par
\end{quote}

The conditional-response analysis uses an AND question about two selected edges. The example below uses the same system message and requires \texttt{B} (\emph{False}), since the first selected edge is absent. The paired context changes only \texttt{Netiralu -> Pisigoli: present.} to \texttt{Netiralu -> Pisigoli: absent.} Both contexts require \emph{False} before intervention, but the answer depends on the first selected edge only when the second is present.

\noindent\textbf{User message}
\begin{quote}
\ttfamily\raggedright
\noindent\detokenize{The directed graph has the following explicit edge facts.}\par
\noindent\detokenize{Netiralu -> Pefumoka: present.}\par
\noindent\detokenize{Pefumoka -> Pisigoli: absent.}\par
\noindent\detokenize{Netiralu -> Pisigoli: present.}\par
\noindent\detokenize{Selected pair:}\par
\noindent\detokenize{First edge: Pefumoka -> Pisigoli.}\par
\noindent\detokenize{Second edge: Netiralu -> Pisigoli.}\par
\noindent\detokenize{Are both selected edges present?}\par
\noindent\detokenize{Response key: A means true; B means false. Reply with exactly A or B. Do not explain.}\par
\end{quote}

Across the eight graphs, each fact-reading task requires four \emph{True} and four \emph{False} answers. \emph{Compose path} and the AND task require \emph{True} on two graphs and \emph{False} on six. Both response keys appear for every graph and instruction, so the required \texttt{A} and \texttt{B} outputs are equally frequent despite the imbalance in truth values.

\subsection{Model checkpoints and input formatting}

Each model uses a fixed checkpoint revision and its accompanying tokenizer. Gemma receives the system instruction and graph prompt in a single user message, while Qwen, Llama, and both Falcon-H1 variants receive separate system and user messages. Qwen uses \texttt{enable\_thinking=False} to disable thinking mode. Answer effects use the first answer token's logits, with \texttt{Yes}/\texttt{No} for the influence--importance analysis and \texttt{A}/\texttt{B} for the remaining analyses. Each label occupies one token at the answer position. For coded answers, the \emph{True}-minus-\emph{False} margin follows the response key, so a positive margin favors \emph{True} under either mapping.

\subsection{Calibration and evaluation sets}

For the influence and importance analyses, four vocabulary groups per model supply the reference activations for directions, perturbation scales, and mean replacement states. Figure~\ref{fig:influence} uses eight additional groups for Falcon-H1-1.5B and 16 for each remaining model. Each evaluation group contributes eight opposite-answer prompts, four per instruction. The influence--importance comparison uses the same 128 prompts for both measurements within each of its four models.

The factual-intervention analysis uses the same 12 vocabulary groups for Gemma and Qwen (Figure~\ref{fig:factual_sources}). Each group contains 192 prompts across eight graphs, two instructions, two response keys, three assignments of node names, and two fact wordings. Effects are averaged across the two wordings within each group before estimating uncertainty.

The internal-edit and response-prediction analyses share 16 evaluation vocabulary groups in Gemma and Qwen (Figures~\ref{fig:task_edit}A, \ref{fig:task_edit}C, \ref{fig:task_edit}D, and \ref{fig:prediction_conditional}A). Each group contains 64 prompts covering all eight graphs, both fact-reading instructions, both response keys, and two statement orders. Four separate groups provide reference activations for the task vector and factual directions. Head restoration uses the 32 opposite-answer prompts within each evaluation group. The all-head analysis uses separate sets of 16 additional groups for each model (Figure~\ref{fig:task_edit}B). Each group contains 32 prompts covering all eight graphs, both fact-reading instructions, and both response keys in a single statement order. The analysis includes all 672 Gemma heads and 1,600 Qwen heads.

The conditional-response analysis uses 16 additional vocabulary groups shared by the fine-tuned Gemma and Qwen models (Figure~\ref{fig:prediction_conditional}B). The AND instruction refers to $B\to C$ and $A\to C$, with each edge targeted in turn. Each group contributes 32 paired comparisons covering both targeted edges, two wordings, both response keys, two statement orders, and both states of the remaining edge.

\subsection{Baseline task performance}

For Figure~\ref{fig:influence}, a baseline answer is correct when the required label has a higher logit than the alternative. Gemma answers 119/128 prompts correctly, Qwen 125/128, and Llama 116/128. The Falcon-H1 models score 57/64 for the 1.5B variant and 128/128 for the 7B variant. All prompts remain in the intervention analyses, including those with incorrect baseline answers.

For Figure~\ref{fig:factual_sources}, correctness requires the highest-logit token to match the specified answer label. Across both instructions and fact wordings, Gemma answers 2,284/2,304 prompts correctly (99.1\%) and Qwen 2,257/2,304 (98.0\%). Both models answer all 1,024 fact-reading prompts correctly before the interventions in Figures~\ref{fig:task_edit}A, \ref{fig:task_edit}C, \ref{fig:task_edit}D, and \ref{fig:prediction_conditional}A. Each model also answers all 512 prompts correctly in the separate all-head evaluation (Figure~\ref{fig:task_edit}B).

Figure~\ref{fig:prediction_conditional}B contains 768 distinct baseline prompts per model, all requiring \emph{False}. Fine-tuned Gemma answers 767/768 correctly (99.9\%), and Qwen 765/768 (99.6\%). Each prompt is counted once, including prompts used in comparisons for both targeted edges.

A separate validation set covers all eight graph configurations for AND on $B\to C$ and $A\to C$, using four additional vocabulary groups. Gemma answers all 64 \emph{True} and 192 \emph{False} prompts correctly. Qwen answers 62/64 \emph{True} and 192/192 \emph{False} prompts correctly. These counts pool both fact wordings, response keys, and statement orders.

\section{Influence and component importance}

\subsection{Component locations and selection}

H1--H3 label attention heads in order of increasing depth. Each head is measured immediately before the attention output projection, and the residual-stream site is measured at the input to a later decoder layer. All sites use the final input position, with layer and head indices starting at zero. The selected sites are:
\begin{itemize}
\item \textbf{Gemma-2-9B-it:} heads at (layer 1, head 6), (layer 26, head 9), and (layer 32, head 7). The residual site is at layer 34.
\item \textbf{Qwen3-14B:} heads at (layer 1, head 1), (layer 24, head 17), and (layer 32, head 13). The residual site is at layer 34.
\item \textbf{Llama-3.1-8B-Instruct:} heads at (layer 0, head 7), (layer 20, head 13), and (layer 24, head 27). The residual site is at layer 28.
\item \textbf{Falcon-H1-1.5B-Deep-Instruct:} heads at (layer 26, head 0), (layer 40, head 4), and (layer 48, head 4). The residual site is at layer 52.
\item \textbf{Falcon-H1-7B-Instruct:} heads at (layer 1, head 8), (layer 28, head 6), and (layer 34, head 3). The residual site is at layer 39.
\end{itemize}

For Gemma, Qwen, Llama, and Falcon-H1-7B, candidate selection divides the model's layers into three depth ranges. Within each range, heads are ranked by the standardized separation between reference activations for prompts requiring \emph{True} and \emph{False} answers. The direction is fitted on three vocabulary groups and evaluated on the fourth, with each group omitted in turn. The four highest-ranked heads from each range form the candidate set. Candidate trios contain one head from each range and span at least one quarter of the model's layer-index range.

Selection evaluates the effect of removing the activation projection along each head's reference answer direction. Two random perturbations provide controls with the same magnitude and directions orthogonal to the answer direction. For a set of heads $S$, damage $D(S)$ is the mean correct-answer margin under the controls minus the margin under projection removal. Head $i$ contributes
\[
C_i(S)=D(S)-D(S_{-i}),
\]
where $S_{-i}$ contains the other two heads. An eligible trio has positive damage in all four vocabulary groups. At least two heads must have positive mean additional contributions, with positive contributions in at least three groups. The second-largest contribution is calculated within each group and then averaged across groups. The trio with the largest mean is selected, with mean damage resolving ties. This criterion favors trios with more than one contributing head.

Residual-site candidates lie two to eight layers after H3, bounded by the final decoder layer. The selected site has the largest standardized separation between task instructions, evaluated by leaving out one reference vocabulary group at a time. Its task direction removes the pooled answer direction, as defined below.

\subsection{Reference directions and activation scales}

Reference activations define a common direction and scale at each site under both instructions. For site $i$, $\mu_{i,t,y}$ is the mean activation under task $t$ with correct semantic answer $y$. The initial task direction averages the instruction difference separately for the two answer classes:
\[
r_i=
\frac{1}{2}
\sum_{y\in\{\mathrm{False},\mathrm{True}\}}
\left(
\mu_{i,\mathrm{direct},y}
-
\mu_{i,\mathrm{path},y}
\right).
\]
The pooled \emph{True}-minus-\emph{False} mean activation difference defines a unit answer direction $\hat a_i$. The normalized task direction is
\[
u_i=
\frac{r_i-(r_i^\top\hat a_i)\hat a_i}
{\left\|r_i-(r_i^\top\hat a_i)\hat a_i\right\|_2}.
\]
This adjustment removes the component along one linear answer direction. Other forms of answer information can remain in the activations.

The reference mean $\bar h_i$ centers each activation, and the sample standard deviation $s_i$ of its reference projections sets the scale. The standard deviation uses denominator $N-1$ for $N$ reference prompts. The resulting coordinate is
\[
z_i(x)=
\frac{u_i^\top\left(h_i(x)-\bar h_i\right)}{s_i}.
\]
One unit corresponds to one standard deviation of the reference projections. The directions, means, and scales remain the same under both instructions.

\subsection{Interventions and task profiles}

The source interventions are $h_i^\pm(x)=h_i(x)\pm2s_i u_i$, applied only to the selected head at the final input position. Influence is calculated from the recipient coordinates using the definition in Section~2.2. The denominator of four reflects the separation between the positive and negative source shifts. The sign indicates whether a positive source shift moves the recipient along or against its task direction. All intervening computation remains active, so the response includes transmission through other components. The measurement concerns a finite perturbation range rather than an infinitesimal derivative or an isolated direct connection.

Importance replaces the selected head's full output vector, or the full residual vector, with $\bar h_i$ and allows the remaining activations to respond. The margin decrease follows Section~2.2. Negative values are retained when replacement increases the correct-answer margin. Within each task, effects are averaged over prompts within a vocabulary group and then equally across groups. The six influence entries are ordered as H1$\to$H2, H1$\to$H3, H1$\to$Res., H2$\to$H3, H2$\to$Res., and H3$\to$Res. The four importance entries follow H1, H2, H3, and Res. Both measurements use identical evaluation prompts within each model.

For either profile $\mathbf v$, the angle between task means is
\[
\theta(\mathbf v)=
\arccos\!\left(
\frac{\mathbf v_{\mathrm{path}}^\top\mathbf v_{\mathrm{direct}}}
{\|\mathbf v_{\mathrm{path}}\|_2\,\|\mathbf v_{\mathrm{direct}}\|_2}
\right).
\]

The profile angle compares relative patterns rather than absolute magnitudes. A uniform positive rescaling leaves the angle unchanged, so a small angle does not require similar absolute effects. Figure~\ref{fig:influence}B instead shows the between-task influence difference normalized within each model:
\[
\widehat{\Delta\mathbf I}
=
\frac{\mathbf I_{\mathrm{direct}}-\mathbf I_{\mathrm{path}}}
{\left\|\mathbf I_{\mathrm{direct}}-\mathbf I_{\mathrm{path}}\right\|_2}.
\]
The normalization shows which site pairs account for the change within a model.

\subsection{Contribution of the residual site to profile orientation}

The later residual site has a larger mean replacement effect than any selected head. Across the four models in Figure~\ref{fig:influence}C and both instructions, its squared effect accounts for 97.8--99.95\% of the importance profile's squared norm. This proportion describes the measured four-site vector, rather than a fraction of the model's total computation. In Gemma, the residual replacement effect is $8.04$ $[7.80,\,8.28]$ logits under \emph{Compose path} and $3.31$ $[2.54,\,4.01]$ under \emph{Read direct edge}. The paired decrease is $4.73$ $[3.98,\,5.60]$ logits, with an interval that excludes zero. The four-site importance angle is $4.15^\circ$ $[1.51^\circ,\,8.18^\circ]$. All intervals have approximate simultaneous 95\% coverage. The small angle reflects the residual site's continued dominance despite the change in effect size.

The exploratory head-only comparison excludes the residual site from both profiles. The influence profile contains H1$\to$H2, H1$\to$H3, and H2$\to$H3, while the importance profile contains the replacement effects for H1, H2, and H3. The influence and importance angles are $0.48^\circ$ and $1.17^\circ$ in Gemma, $30.91^\circ$ and $9.42^\circ$ in Qwen, $13.48^\circ$ and $4.89^\circ$ in Llama, and $8.50^\circ$ and $12.65^\circ$ in Falcon-H1-7B. These are descriptive point estimates; all four pointwise 95\% paired bootstrap intervals for the angle differences include zero. Figure~\ref{fig:influence}C uses the full profiles, with six influence values and four importance values including the residual site.

\section{Factual interventions and downstream restoration}

\subsection{Factual activation differences}

The factual interventions in Figure~\ref{fig:factual_sources} act at the input to decoder layer 12 in both Gemma and Qwen. The changed residual activations span the complete fact statement, including the node names, relation, and presence or absence statement. A reference pair differs only in the targeted fact, for example:
\begin{quote}
\ttfamily\raggedright
\noindent\detokenize{Pibepaka -> Kupepezi: present.}\par
\noindent\detokenize{Pibepaka -> Kupepezi: absent.}\par
\end{quote}
Both reference prompts use \emph{Compose path} with \texttt{A} for \emph{True} and \texttt{B} for \emph{False}. The other facts, node names, and statement order remain identical. Each graph and wording has its own reference pair, with aligned statement positions.

For fact $f$, $H_f^{\mathrm{ref}}$ and $H_f^{\mathrm{flip}}$ contain the residual vectors across the original and changed statements. Their difference supplies the intervention:
\[
\Delta_f=H_f^{\mathrm{flip}}-H_f^{\mathrm{ref}},
\qquad
H_{f,t}'=H_{f,t}+\Delta_f,
\]
where $t$ denotes the task instruction. The same displacement is applied under both instructions and response keys while the written prompt remains fixed. Two source controls each combine a random permutation of the displacement's dimensions with random sign flips. Each transformation preserves the Euclidean magnitude at every intervened position.

\subsection{Answer effects and task selectivity}

The answer score $q_t$ is the logit for the label representing \emph{True} minus the logit for the label representing \emph{False}. Its orientation follows the response key when the meanings of \texttt{A} and \texttt{B} are exchanged. For source variant $v$, the signed answer effect is
\[
E_{f,t}^{(v)}
=
\sigma_f\left(q_{f,t}^{(v)}-q_t^{\mathrm{base}}\right),
\]
where $\sigma_f=+1$ for an absent-to-present intervention and $-1$ for a present-to-absent intervention. The superscript $\mathrm{base}$ denotes the unaltered run, and $v$ denotes the factual displacement or either source control. Positive effects favor \emph{True} for an absent-to-present intervention and \emph{False} for a present-to-absent intervention. The adjusted effect subtracts the mean source-control effect:
\[
\widetilde E_{f,t}
=
E_{f,t}^{(\mathrm{fact})}
-
\frac{E_{f,t}^{(\mathrm{control\,1})}
      +E_{f,t}^{(\mathrm{control\,2})}}{2}.
\]

Task selectivity compares the adjusted effects under the relevant and irrelevant instructions:
\[
T_f
=
\left\langle\widetilde E_{f,\mathrm{relevant}}\right\rangle
-
\left\langle\widetilde E_{f,\mathrm{irrelevant}}\right\rangle.
\]
For $A\to B$, both task conditions use graphs with $B\to C$ present. The \emph{Compose path} answer then depends on $A\to B$, while the \emph{Read direct edge} answer is independent of that edge. The $B\to C$ intervention uses the same comparison with $A\to B$ present. For $A\to C$, \emph{Read direct edge} is relevant and \emph{Compose path} is irrelevant across all graph states.

\subsection{Later activation restoration}

Downstream restoration replaces later residual activations with their values from an unaltered run of the same prompt. The factual intervention or source control remains active at layer 12. The restoration layer is 23 in Gemma and 32 in Qwen. The restored positions cover either the complete task question or the response-key span. The latter extends from the end of the question through the final input position, including the answer mapping and response instructions. Both spans are disjoint from the graph-fact statements. Figure~\ref{fig:factual_sources}B uses the response-key span.

For source variant $v$, the restoration displacement is
\[
\Delta_{\mathrm{restore}}^{(v)}
=
H_{\mathrm{later}}^{\mathrm{base}}
-
H_{\mathrm{later}}^{(v)}.
\]
Two restoration controls each permute the dimensions of this displacement and randomly flip their signs. For each source variant, we apply the original restoration displacement and its two randomized versions to the corresponding later state in separate runs.

\begin{figure}[!htb]
    \centering
    \includegraphics[width=\linewidth]{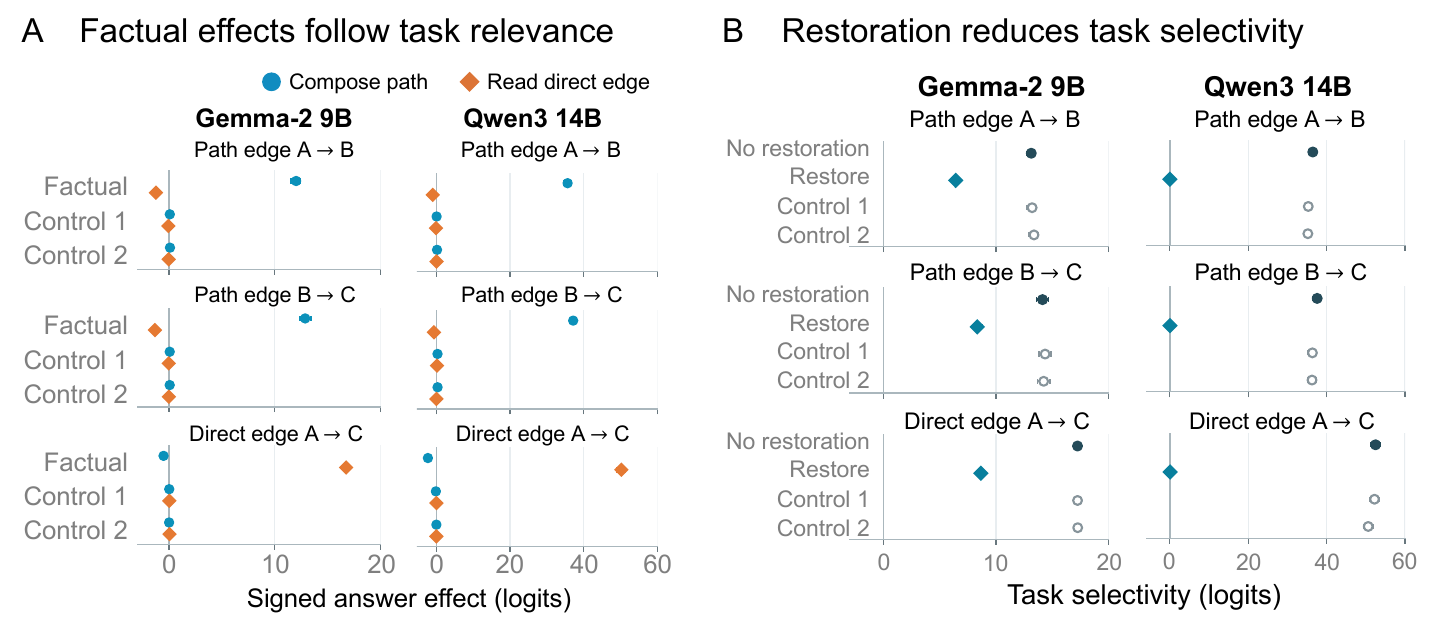}
    \caption{\textbf{Source controls and downstream restoration.} \textbf{A. Factual effects follow task relevance.} Points show signed answer effects of factual interventions and two source controls under \emph{Compose path} and \emph{Read direct edge}, before source-control subtraction. Positive effects favor \emph{True} for an absent-to-present intervention and \emph{False} for a present-to-absent intervention. \textbf{B. Restoration reduces task selectivity.} Task selectivity is shown without a downstream intervention, under restoration of the original response-key activations, and under two restoration controls. Every estimate in panel B includes the source-control adjustment, recalculated for that downstream condition. The factual intervention remains active during restoration. Each source or restoration control combines a random permutation of the displacement's dimensions with random sign flips, preserving its magnitude at each intervened position. Both panels use the models, prompts, and intervention sites from Figure~\ref{fig:factual_sources}. Points show means over 12 vocabulary groups, with both wordings averaged within each group. Whiskers show pointwise 95\% confidence intervals from 10,000 paired bootstrap resamples of vocabulary groups.}
    \label{fig:supp_factual_controls}
\end{figure}

Task selectivity is recalculated under each downstream condition, retaining the source-control adjustment in Section~C.2. The quantities $T_f^{\mathrm{intact}}$, $T_f^{\mathrm{restore}}$, and $T_f^{(k)}$ denote selectivity without a downstream intervention, under restoration, and under restoration control $k$. The controlled contribution is
\[
D_f=
\left(T_f^{\mathrm{intact}}-T_f^{\mathrm{restore}}\right)
-
\frac{1}{2}\sum_{k=1}^{2}
\left(T_f^{\mathrm{intact}}-T_f^{(k)}\right).
\]
For Gemma's $A\to B$ intervention, task selectivity is $13.14$ $[12.72,\,13.52]$ logits without a downstream intervention. Selectivity falls to $6.43$ $[6.06,\,6.83]$ under restoration and averages $13.30$ $[12.88,\,13.68]$ under the restoration controls. The three intervals are pointwise 95\% confidence intervals. The difference between the mean control and restoration values gives a controlled contribution of $6.87$ $[6.28,\,7.39]$ logits. Its corrected 95\% confidence interval excludes zero. This effect measures the reduction beyond the restoration controls and includes transmission through intervening model components. Figure~\ref{fig:supp_factual_controls} shows the factual effects and restoration comparisons alongside their controls for all three targeted facts.

\section{Internal task edits and factual responses}

\subsection{Construction of the task edit}

The internal edit changes residual activations before layer 12 in Gemma and Qwen, from the instruction position specifying which fact to read through the final input position. For example, the paired questions differ only in the number in \textquotedblleft{}Is the edge in Fact 1 present?\textquotedblright{} and \textquotedblleft{}Is the edge in Fact 2 present?\textquotedblright{} Reference pairs require the same answer under both instructions. Facts 1 and 2 are therefore either both present or both absent, while Fact 3 takes both possible values. The mean activation difference is
\[
\Delta_{\mathrm{task}}
=
\left\langle
H_{\mathrm{Read\ fact\ 2}}-H_{\mathrm{Read\ fact\ 1}}
\right\rangle,
\]
where $H$ contains the residual vectors across the edited positions. The average gives equal weight to vocabulary groups, statement orders, response keys, and answer classes. The edit adds $\Delta_{\mathrm{task}}$ under \emph{Read fact 1} and subtracts it under \emph{Read fact 2}, without additional scaling. The displacement remains fixed across changes to the graph facts.

Two random early edits match both the displacement magnitude and the resulting-state magnitude. The residual vectors across the edited positions are concatenated for this calculation. Each control preserves the component of the task displacement parallel to the original state and randomizes the orthogonal component while preserving its length. The matching applies across the complete edited span.

\subsection{Responses to individual facts}

\begin{figure}[!htb]
    \centering
    \includegraphics[width=\linewidth]{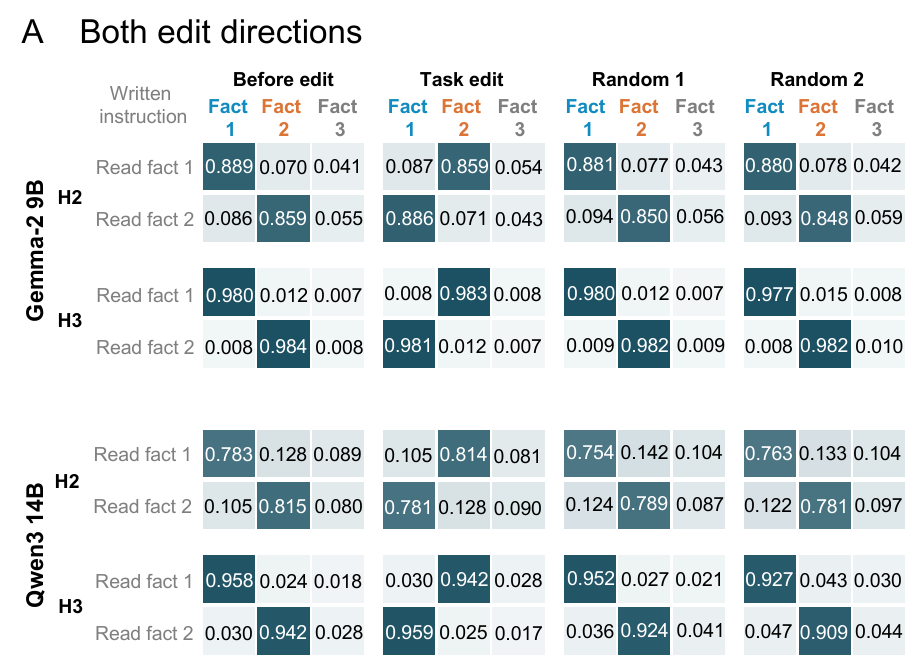}
    \caption{\textbf{Factual responses under both internal edit directions.} Cells show each graph fact's share of squared head-output responses. Each share uses the sum over all three facts as its denominator and is calculated within a vocabulary group and written instruction before averaging over the 16 groups from Figure~\ref{fig:task_edit}A. The task edit targets Fact 2 under \emph{Read fact 1} and Fact 1 under \emph{Read fact 2}, while the written instruction remains fixed. The two random early edits match the task edit's displacement magnitude and resulting-state magnitude. The same H2 and H3 heads are used throughout: H26.09 and H32.07 in Gemma, and H24.17 and H32.13 in Qwen, with zero-based layer and head indices.}
    \label{fig:supp_factual_shares}
\end{figure}

Factual responses compare head outputs for paired prompts that differ in the presence or absence of one graph fact. The instruction, response key, statement order, and other facts remain fixed within each pair. Both prompts receive the same task displacement under the edited condition. For head $i$, fact $f$, and condition $e$, response energy is
\[
E_{i,f}^{(e)}
=
\left\langle
\left\|h_i^{(e)}(x_{f=1})-h_i^{(e)}(x_{f=0})\right\|_2^2
\right\rangle.
\]
Outputs are measured before the attention output projection at the final input position. The average includes factual pairs within a vocabulary group and written instruction. The normalized share follows Section~2.4, with all three facts in the denominator and zero shares for an exactly zero denominator. Shares are calculated within each vocabulary group before averaging across groups. Figure~\ref{fig:task_edit}A shows H2 and H3 under \emph{Read fact 1}. The displayed shares for Facts 1 and 2 need not sum to one because Fact 3 also contributes. The shares describe relative sensitivity, while the unnormalized energies measure the absolute response to each fact. Figure~\ref{fig:supp_factual_shares} extends the fixed-head comparison to all three facts, both edit directions, and both random early edits.

Table~\ref{tab:supp_factual_energies} reports absolute responses before and after the task edit under \emph{Read fact 1}, averaged over 16 vocabulary groups.

\begin{table}[htbp]
    \centering
    \caption{\textbf{Absolute factual-response energies.} Entries show unedited $\to$ task-edited mean squared head-output responses under \emph{Read fact 1}, averaged over 16 vocabulary groups.}
    \label{tab:supp_factual_energies}
    \begin{tabular}{llccc}
        \hline
        Model & Head & Fact 1 & Fact 2 & Fact 3 \\
        \hline
        Gemma & H2 & $81.638 \to 8.427$ & $6.459 \to 83.450$ & $3.773 \to 5.242$ \\
        Gemma & H3 & $16.421 \to 0.141$ & $0.208 \to 16.674$ & $0.120 \to 0.138$ \\
        Qwen & H2 & $23.356 \to 2.930$ & $3.812 \to 22.692$ & $2.665 \to 2.250$ \\
        Qwen & H3 & $63.412 \to 2.071$ & $1.556 \to 66.308$ & $1.138 \to 1.942$ \\
        \hline
    \end{tabular}
\end{table}

The all-head analysis uses the reassignment score $Q$ defined in Section~2.4, calculated separately for each head. Positive scores indicate a shift toward the alternative fact. The score lies between $-1$ and $1$, with zero assigned when the denominator is zero. Response energies pool both response keys before normalization, and each head's score is then averaged across both edit directions and vocabulary groups. Figure~\ref{fig:task_edit}B shows these head means and their median within each layer. The plotted spread represents variation across heads. All 672 Gemma heads and 1,600 Qwen heads are included.

The population comparison divides post-edit layers into three bands: 13--22, 23--32, and 33--41 in Gemma, and 13--21, 22--30, and 31--39 in Qwen. Each band's statistic is the median of its heads' mean reassignment scores. The control-adjusted score subtracts the mean score under the two random early edits.

\subsection{Influence before and after the edit}

The influence analysis uses the four Gemma and Qwen sites in Appendix~B, with directions derived from factual activation differences. Under \emph{Read fact 1}, the reference direction at each site is the mean activation difference between Fact 1 being present and absent. The second direction uses the corresponding difference for Fact 2 under \emph{Read fact 2}. For factual direction $k$, influence from source $i$ to recipient $j$ is
\[
I_{ij}^{(k)}
=
\frac{
\left(u_j^{(k)}\right)^\top
\left(h_j^{+,k}-h_j^{-,k}\right)
}{4s_j^{(k)}}.
\]
Here, $h_j^{+,k}$ and $h_j^{-,k}$ are the recipient outputs after positive and negative source shifts along factual direction $k$. The source shifts use the standardized two-unit procedure in Section~2.2. The recipient direction and scale are $u_j^{(k)}$ and $s_j^{(k)}$. The two directional estimates are averaged for each of the six forward site pairs. All intervening computation remains active. The early edit is added to the ongoing residual state, allowing an H1 perturbation to propagate into the layer-12 intervention.

Within each vocabulary group and written instruction, the mean six-pair profile is compared with the unedited profile under the alternative instruction:
\[
d_e=
\left\|\overline{\mathbf I}_e
-\overline{\mathbf I}_{\mathrm{alternative}}\right\|_2^2.
\]
The target prompts have matching graph facts, response keys, and statement orders. Each profile is averaged before its squared distance is calculated, and the two edit-direction distances are then averaged within each vocabulary group. The random-edit comparison averages the distances for the two controls after calculating each separately. Figure~\ref{fig:task_edit}C reports the reduction as $100(1-\langle d_{\mathrm{task}}\rangle/\langle d_{\mathrm{unedited}}\rangle)$, with equal weight to vocabulary groups in each mean.

\subsection{Individual head contributions}

Head restoration replaces H2 or H3 with its output from the same prompt without an early edit. The early intervention remains active, and the other head remains free to respond. The comparison uses opposite-answer graphs, for which the written and alternative instructions require opposite answers. For head $i$ under early condition $e$, the restoration effect is
\[
R_{i,e}=m_{i,e}^{\mathrm{restore}}-m_e,
\]
where $m_e$ is the logit margin for the answer required by the written instruction. A positive effect moves the margin toward that answer. Two random head changes match both the restoration-displacement magnitude and the resulting-output magnitude, using the construction in Section~D.1. The adjusted effect is
\[
A_{i,e}
=
R_{i,e}-\frac{C_{i,e}^{(1)}+C_{i,e}^{(2)}}{2},
\]
where $C_{i,e}^{(1)}$ and $C_{i,e}^{(2)}$ are the answer-margin changes under the head controls.

The adjustment is calculated under the task edit and each random early edit. The controlled contribution in Figure~\ref{fig:task_edit}D is
\[
K_i
=
A_{i,\mathrm{task}}
-
\frac{
A_{i,\mathrm{early\ control\ 1}}
+A_{i,\mathrm{early\ control\ 2}}
}{2}.
\]
This comparison accounts for random changes at both the early intervention site and the restored head. Effects are averaged within each vocabulary group and edit direction, followed by equal averages across directions and groups. On the same opposite-answer prompts, Gemma's mean margin for the written instruction's answer changes from $+8.99$ $[8.92,\,9.05]$ to $-8.97$ $[-9.04,\,-8.90]$ logits after the task edit. Qwen's mean margin changes from $+24.10$ $[23.94,\,24.25]$ to $-24.13$ $[-24.29,\,-23.97]$ logits. The paired shifts toward the alternative answer are $17.96$ $[17.82,\,18.09]$ logits in Gemma and $48.23$ $[47.91,\,48.54]$ in Qwen. All intervals have approximate simultaneous 95\% coverage, and both shift intervals exclude zero.
\section{Prediction of downstream factual responses}

\subsection{Response maps and factual interventions}

The response map in Section~2.5 uses the full H2 and H3 output differences under positive and negative perturbations along two factual directions. Each prompt provides an unedited map and a task-edited map, denoted $M_e$ for internal condition $e$. The selected heads are the H2 and H3 used in Figure~\ref{fig:task_edit}. A directional term contributes a zero map when its measured H2 difference is exactly zero.

The separate factual intervention replaces H2's output with its value from a reference prompt in which one graph fact has changed. The instruction, response key, statement order, and other facts remain identical. Under the edited condition, the original and reference runs receive the same early task displacement. The measured H2 displacement $\mathbf d_{f,e}$ and resulting H3 change $\mathbf y_{f,e}$ are relative to the original prompt under the same internal condition. The two predictions are
\[
\begin{array}{rcl}
\widehat{\mathbf y}_{f,e}^{\,\mathrm{specific}}
&=&M_e\mathbf d_{f,e},\\
\widehat{\mathbf y}_{f,e}^{\,\mathrm{averaged}}
&=&\displaystyle\frac{M_{\mathrm{unedited}}+M_{\mathrm{edited}}}{2}\mathbf d_{f,e}.
\end{array}
\]
Both predictors receive the same measured H2 displacement. H3's response to the factual intervention enters only the error calculation, while the maps use H3 responses to the controlled perturbations. Prediction therefore requires the controlled-perturbation measurements and factual H2 displacement for the evaluated prompt.

\subsection{Prediction errors and figure normalization}

For predictor $p\in\{\mathrm{specific},\mathrm{averaged}\}$, the error includes every dimension of the H3 output:
\[
L_p=\left\|\mathbf y_{f,e}-\widehat{\mathbf y}_{f,e}^{\,p}\right\|_2^2.
\]
A prediction of no H3 change has error $L_{\mathrm{zero}}=\|\mathbf y_{f,e}\|_2^2$. Within each vocabulary group, errors are averaged over all three factual interventions, both written instructions, both response keys, both statement orders, and all graph states. The unedited and edited conditions receive equal weight, followed by an equal average across the 16 vocabulary groups. Figure~\ref{fig:prediction_conditional}A divides every group's error by the model's mean $L_{\mathrm{zero}}$. Both predictors share this denominator across all groups, and the dashed line at one marks the mean error of predicting no response.

The percentage reduction is $100(1-\langle L_{\mathrm{specific}}\rangle/\langle L_{\mathrm{averaged}}\rangle)$, with errors averaged before division. In Gemma, the task-specific, averaged-map, and no-response mean errors are 0.00747 $[0.00725,\,0.00770]$, 0.01610 $[0.01572,\,0.01648]$, and 0.04076 $[0.03958,\,0.04184]$, respectively. In Qwen, the corresponding errors are 0.09999 $[0.09700,\,0.10324]$, 0.10519 $[0.10193,\,0.10876]$, and 0.12402 $[0.12021,\,0.12852]$. All six intervals are pointwise 95\% confidence intervals.

The task-specific map reduces mean error relative to the averaged map by 53.6\% $[52.2\%,\,54.9\%]$ in Gemma and 5.0\% $[4.0\%,\,5.9\%]$ in Qwen, with approximate simultaneous 95\% intervals. The paired absolute reductions are 0.00863 $[0.00817,\,0.00909]$ and 0.00521 $[0.00379,\,0.00663]$, respectively, retaining the four-comparison correction in Appendix~G. Both corrected difference intervals exclude zero.

Separate averages for each fact show that the improvement is concentrated in the two facts requested by the reading instructions, both before and after the internal edit. For the third fact, the absolute error differences are below 0.00022, with Qwen's unedited mean favoring the averaged map. These fact-wise summaries are descriptive.

\subsection{Prediction in fine-tuned models}

The same prediction analysis uses the Gemma and Qwen checkpoints fine-tuned on graph tasks for the conditional-response experiment in Figure~\ref{fig:prediction_conditional}B. The H2--H3 pairs remain unchanged, while the early task displacement and factual directions are estimated from reference activations in each fine-tuned checkpoint. In Gemma, mean squared error decreases from 0.01549 $[0.01495,\,0.01602]$ with the averaged map to 0.00760 $[0.00721,\,0.00798]$ with the task-specific map. In Qwen, the corresponding errors are 0.08742 $[0.08451,\,0.09060]$ and 0.08548 $[0.08277,\,0.08838]$. All four intervals are pointwise 95\% confidence intervals.

The percentage reductions are 50.9\% $[49.6\%,\,52.4\%]$ in Gemma and 2.2\% $[1.4\%,\,3.1\%]$ in Qwen, with approximate simultaneous 95\% intervals. The paired absolute reductions are 0.00789 $[0.00743,\,0.00834]$ and 0.00194 $[0.00075,\,0.00314]$, respectively, retaining the 12-comparison correction in Appendix~G. Both corrected difference intervals exclude zero.

The mean errors for predicting no H3 change are 0.03991 $[0.03891,\,0.04089]$ in Gemma and 0.10061 $[0.09761,\,0.10410]$ in Qwen, with pointwise 95\% confidence intervals. The task-specific map reduces error relative to this baseline by 0.03231 $[0.03105,\,0.03357]$ in Gemma and 0.01513 $[0.01287,\,0.01739]$ in Qwen. These simultaneous 95\% intervals retain the 12-comparison correction in Appendix~G. Both corrected difference intervals exclude zero.
\section{Conditional responses in fine-tuned models}

\subsection{Fine-tuning and checkpoint selection}

The training curriculum combines four rules---read the first selected edge, read the second, AND, and OR---with two edge pairs: $(A\to B,B\to C)$ and $(B\to C,A\to C)$. AND with the second pair and OR with the first are omitted from training, leaving six rule--pair combinations. Each combination includes all eight graph states, two fact wordings, two response keys, and two prompt layouts. Across 32 vocabulary groups, the training set contains 12,288 prompts. Sampling weights balance the six combinations and the two truth labels within each combination.

Both models use rank-8 LoRA adapters \citep{hu2022lora} on the attention query and value projections, with scaling parameter 16 and no adapter dropout. The base weights remain frozen in bfloat16, while the adapter parameters use float32. Training minimizes cross-entropy for the first answer token over the full output vocabulary. AdamW uses no weight decay and an effective batch size of 32. The learning rate increases to $2\times10^{-4}$ over 32 updates, followed by cosine decay scheduled over 512 updates. The gradient norm is clipped at 1. Gemma uses batches of 32, while Qwen accumulates gradients over eight batches of four.

Gemma uses a 512-update checkpoint selected during earlier development from nine candidates across three training runs. That selection included rule--pair combinations omitted from training. Qwen uses the first scheduled checkpoint meeting its behavioral criteria, at 128 updates. Its selection uses four development vocabulary groups containing only the trained combinations. For each wording and response key, qualification requires mean balanced accuracy of at least 90\%, accuracy of at least 85\% in every rule--pair and truth-label group, and at least 99\% valid \texttt{A}/\texttt{B} outputs.

\subsection{Fixed displacements across factual contexts}

Figure~\ref{fig:prediction_conditional}B uses the AND instruction for $B\to C$ and $A\to C$, with each edge targeted in turn. The targeted edge is initially absent in both contexts. The other selected edge is present in one context and absent in the other, so both initial answers are \emph{False}, but the targeted fact determines the answer only in the first context. The H2 displacement is estimated with the other selected edge absent:
\[
\mathbf d_f
=
\mathbf h_2(f=1,g=0)
-
\mathbf h_2(f=0,g=0),
\]
where $f$ and $g$ indicate the presence of the targeted and other selected edges. The reference pair preserves the remaining fact, instruction, response key, and prompt layout. Both contexts receive the same displacement $\mathbf d_f$ at H2. The H3 response is the full output difference from the unaltered run in the same context. Response strength is the squared norm of this difference. The conditional effect is response strength when the targeted fact determines the answer minus response strength when it does not.

One source control randomly permutes the displacement's dimensions, and the other reverses the signs of a randomly selected half of its dimensions. Both preserve the displacement's norm. Each control undergoes the same comparison between factual contexts. The adjusted conditional effect subtracts the mean of these two control effects from the factual displacement's effect. Effects are averaged within each vocabulary group over targeted edges, wordings, response keys, layouts, and the remaining edge's state, followed by an equal average across groups. The relevant-minus-irrelevant difference in squared H3 response is 0.01248 $[0.00841,\,0.01655]$ in Gemma and 0.03908 $[0.02115,\,0.05700]$ in Qwen. The excess over the mean difference produced by the permutation and sign-flip controls is 0.01347 $[0.00731,\,0.01963]$ in Gemma and 0.02320 $[0.01323,\,0.03317]$ in Qwen. All four simultaneous 95\% intervals exclude zero under the 12-comparison correction in Appendix~G.

The control-adjusted effects in Gemma are 0.03180 $[0.02282,\,0.04077]$ for $B\to C$ and $-0.00485$ $[-0.00938,\,-0.00032]$ for $A\to C$. Qwen's corresponding effects are 0.02922 $[0.01337,\,0.04508]$ and 0.01717 $[0.00779,\,0.02656]$. The paired differences between facts ($B\to C$ minus $A\to C$) are 0.03665 $[0.02775,\,0.04555]$ in Gemma and 0.01205 $[-0.00684,\,0.03094]$ in Qwen. All six estimates have simultaneous 95\% intervals with Bonferroni correction (Appendix~G). The between-fact interval excludes zero in Gemma and includes zero in Qwen.

\section{Aggregation and statistical uncertainty}

\subsection{Paired averaging and resampling}

Within-group summaries follow the averaging order specified for each analysis, and vocabulary groups receive equal weight. Bootstrap resampling retains all paired conditions within each selected group. The confidence intervals quantify variation across vocabulary groups, conditional on the selected component sites and model checkpoints. Figures~\ref{fig:influence} and~\ref{fig:factual_sources} use pointwise 95\% confidence intervals from 10,000 resamples. Task means and derived quantities, including profile angles, are recalculated in each resample, and the 2.5th and 97.5th percentiles give the interval bounds.

\subsection{Influence, importance, and factual interventions}

For the influence--importance comparison, we calculate the influence angle minus the importance angle within each of 10,000 paired bootstrap resamples of vocabulary groups. Each resample preserves both task instructions and both measurements. Bonferroni correction across four models uses the 0.625th and 99.375th percentiles, giving approximate simultaneous 95\% confidence intervals. The null hypothesis of equal angles is rejected when the corresponding interval excludes zero. The Gemma residual-site example uses 10,000 paired bootstrap resamples of vocabulary groups. Each resample recalculates the two residual replacement means, their paired difference, and the four-site importance angle. Bonferroni correction across these four quantities gives approximate simultaneous 95\% confidence intervals.

The factual-intervention analysis assesses task selectivity and downstream contribution across three facts and two models, giving 12 comparisons. Each of 10,000 bootstrap resamples preserves all paired conditions within each vocabulary group. Bonferroni correction across the 12 comparisons uses the 0.2083rd and 99.7917th percentiles, giving approximate simultaneous 95\% confidence intervals. A zero effect is rejected when the corresponding interval excludes zero. The three selectivity means in the Gemma restoration example use pointwise 95\% bootstrap intervals. The two restoration controls are averaged within each vocabulary group before resampling. The controlled contribution retains the 12-comparison correction used for Figure~\ref{fig:factual_sources}.

\subsection{Internal task edits and head restoration}

The internal-edit analysis assesses 24 comparisons together. Factual reassignment and individual head restoration each contribute eight comparisons across two models, two heads, and raw and control-adjusted effects. Influence-profile movement contributes another eight across two models, two graph subsets, and raw and control-adjusted effects. The graph subsets comprise all graphs and those for which the two instructions require the same answer.

Approximate simultaneous 95\% intervals use 20,000 paired bootstrap resamples of the 16 vocabulary groups. Each resample provides the largest absolute deviation across comparisons, with deviations divided by the corresponding standard errors from the original sample. The 95th percentile of these maxima gives the bootstrap critical value, $c_{\mathrm{boot}}$. The interval for comparison $j$ is
\[
\bar{x}_j
\;\pm\;
\max\!\left\{
c_{\mathrm{boot}},
t_{15}\!\left(1-\frac{0.05}{2K}\right)
\right\}
\mathrm{SE}_j,
\]
where $\bar{x}_j$ is the mean vocabulary-level effect, $\mathrm{SE}_j$ is its sample standard deviation divided by $\sqrt{16}$, and $K=24$. The function $t_{15}(p)$ gives the $p$-quantile of a Student-$t$ distribution with 15 degrees of freedom. The Student-$t$ term supplies a Bonferroni-adjusted minimum interval width.

The four response-share changes in Figure~\ref{fig:task_edit}A use 100,000 paired bootstrap resamples of vocabulary groups, with Bonferroni correction across the four comparisons. The depth-band analysis uses its original 100,000 paired resamples and pointwise 97.5\% intervals. Its prespecified criterion requires positive lower bounds in all three bands for both raw and control-adjusted scores. Heads and edit directions remain together within each resampled vocabulary group.

The before-and-after margin comparison uses 100,000 paired bootstrap resamples of vocabulary groups. Each resample recalculates both margins and their paired decrease in each model. The decrease is the written-answer margin before the edit minus the margin after it. Bonferroni correction across the six estimates gives approximate simultaneous 95\% confidence intervals.

The relative restoration effect is the mean control-adjusted restoration effect divided by the mean edit-induced margin shift, expressed as a percentage. Each of 100,000 paired bootstrap resamples recalculates both means from the same vocabulary groups. Bonferroni correction across the four ratios gives approximate simultaneous 95\% intervals.

\subsection{Response prediction}

The prediction analysis in models without additional training uses the simultaneous-interval procedure in Appendix~G.3 with $K=4$. Its comparisons measure the task-specific predictor's error reduction relative to the averaged map and the no-response predictor in each model. The prediction and conditional-response analyses in fine-tuned models use Student-$t$ intervals with 15 degrees of freedom and a Bonferroni correction across 12 comparisons.

For models without additional training, prediction means and percentage reductions use 100,000 paired bootstrap resamples of vocabulary groups. The three mean errors in each model receive pointwise 95\% intervals. Each resample recalculates the percentage reduction from the two predictor means, with Bonferroni correction across the two model ratios giving approximate simultaneous 95\% intervals. The paired absolute reductions retain the original four-comparison procedure. An interval excluding zero rejects equal mean prediction errors for the corresponding pair of predictors.

The task-specific, averaged-map, and no-response mean errors in the fine-tuned models use the same pointwise bootstrap procedure. Their percentage reductions use 100,000 paired resamples with Bonferroni correction across the two fine-tuned model ratios. The paired absolute reductions retain the 12-comparison Student-$t$ intervals.

\subsection{Conditional responses}

The raw conditional-response test compares mean squared H3 responses in the relevant and irrelevant factual contexts. The adjusted test compares the factual displacement's context difference with the mean context difference under the permutation and sign-flip controls. The raw and adjusted differences retain the 12-comparison Student-$t$ correction described in Appendix~G.4. A corrected interval excluding zero rejects the corresponding null difference.

The fact-specific follow-up uses Student-$t$ intervals with 15 degrees of freedom and Bonferroni correction across six estimates: four control-adjusted fact means and two between-fact differences. Each difference subtracts the $A\to C$ effect from the $B\to C$ effect within the same vocabulary group. An interval excluding zero rejects the corresponding null effect or null difference.

The conditional-response ratio divides the mean squared H3 response in the relevant context by that in the irrelevant context. Each of 100,000 paired bootstrap resamples recalculates both means from the same vocabulary groups. Bonferroni correction across the two model ratios gives approximate simultaneous 95\% intervals.
\end{document}